\documentclass[10pt,twocolumn,letterpaper]{article}
\usepackage{iccv}
\usepackage{times}

\usepackage{graphicx}
\usepackage{amsmath}
\usepackage{amssymb}
\usepackage{booktabs}
\usepackage{comment}
\usepackage{bm}
\usepackage{adjustbox}
\usepackage{array}
\newcolumntype{R}[2]{%
    >{\adjustbox{angle=#1,lap=1.3\width-(#2)}\bgroup}%
    l%
    <{\egroup}%
}
\usepackage{pifont}
\usepackage{stmaryrd}
\usepackage{multirow}
\usepackage{enumitem}
\usepackage{flushend}
\usepackage{makecell}
\usepackage{setspace}
\usepackage{cuted}
\usepackage{colortbl}
\usepackage{titletoc}
\usepackage[pagebackref=true,breaklinks=true,letterpaper=true,colorlinks,bookmarks=false]{hyperref}
\usepackage[dvipsnames]{xcolor}

\usepackage{pgfplots,pgfplotstable}
\usepackage{tikz}
\pgfplotsset{compat=1.10}
\pgfplotsset{every axis label/.append style={font=\large}}
\pgfplotsset{every tick label/.append style={font=\large}}
\usetikzlibrary{intersections}
\usepgfplotslibrary{fillbetween}

\iccvfinalcopy 
\def\iccvPaperID{} 

\usepackage{listings}
\definecolor{codegreen}{rgb}{0,0.6,0}
\definecolor{codegray}{rgb}{0.5,0.5,0.5}
\definecolor{codepurple}{rgb}{0.58,0,0.82}
\definecolor{backcolour}{rgb}{0.95,0.95,0.92}
\lstdefinestyle{mystyle}{
    backgroundcolor=\color{backcolour},   
    commentstyle=\color{codegreen},
    keywordstyle=\color{magenta},
    numberstyle=\tiny\color{codegray},
    stringstyle=\color{codepurple},
    basicstyle=\ttfamily\footnotesize,
    breakatwhitespace=false,         
    breaklines=true,                 
    captionpos=b,                    
    keepspaces=true,                 
    numbers=left,                    
    numbersep=5pt,                  
    showspaces=false,                
    showstringspaces=false,
    showtabs=false,                  
    tabsize=2
}
\usepackage[capitalize]{cleveref}
\crefname{section}{Sec.}{Secs.}
\Crefname{section}{Section}{Sections}
\Crefname{table}{Table}{Tables}
\crefname{table}{Tab.}{Tabs.}
\newcommand{\Rbb}{\ensuremath{\mathbb{R}}}

\renewcommand{\geq}{\ensuremath{\geqslant}}
\newcommand{\adjoint}{\ensuremath{{\intercal}}}
\newcommand{\ma}[1]{\ensuremath{\mathsf{#1}}}
\renewcommand{\vec}[1]{\ensuremath{\bm{#1}}}

\newcommand{\set}[1]{\ensuremath{\mathcal{#1}}}

\renewcommand{\th}{\ensuremath{\text{th}}}
\definecolor{free}{rgb}{0.19, 0.55, 0.91}
\definecolor{blue1}{rgb}{0.227, 0.42, 0.912}
\definecolor{blue2}{rgb}{0.427, 0.62, 0.922}
\definecolor{blue3}{rgb}{0.643, 0.761, 0.957}
\definecolor{blue4}{rgb}{0.788, 0.855, 0.973}
\newcommand{\cmark}{\textcolor{free}{\ding{51}}}
\newcommand{\xmark}{}
\newcommand{\ourlayer}{\ensuremath{{\rm WI}}} 
\def \ours{WaffleIron\xspace} 
\newcommand{\smallparagraph}[1]{\medskip\noindent\textbf{#1}~}

\newcommand{\ra}[1]{\renewcommand{\arraystretch}{#1}}

\begin{document}

\title{Using a Waffle Iron for Automotive Point Cloud Semantic Segmentation}

\author{%
Gilles Puy$^1$
\and
Alexandre Boulch$^1$
\and
Renaud Marlet$^{1,2}$
\and
\large
\hspace{-3mm}\textsuperscript{1}valeo.ai, Paris, France  \hspace{1mm} \textsuperscript{2}LIGM, Ecole des Ponts, Univ Gustave Eiffel, CNRS, Marne-la-Vall\'ee, France
}
\maketitle

\begin{abstract}
Semantic segmentation of point clouds in autonomous driving datasets requires techniques that can process large numbers of points efficiently. Sparse 3D convolutions have become the de-facto tools to construct deep neural networks for this task: they exploit point cloud sparsity to reduce the memory and computational loads and are at the core of today's best methods. In this paper, we propose an alternative method that reaches the level of state-of-the-art methods without requiring sparse convolutions. We actually show that such level of performance is achievable by relying on tools a priori unfit for large scale and high-performing 3D perception. In particular, we propose a novel 3D backbone, \ours, made almost exclusively of MLPs and dense 2D convolutions and present how to train it to reach high performance on SemanticKITTI and nuScenes. We believe that \ours is a compelling alternative to backbones using sparse 3D convolutions, especially in frameworks and on hardware where those convolutions are not readily available. The code is available at \url{https://github.com/valeoai/WaffleIron}.
\end{abstract}

%
\section{Introduction}

\begin{figure*}[t]
\begin{center}
\includegraphics[width=0.8\linewidth,trim={3.35cm 5.8cm 11.75cm 1.9cm},clip]{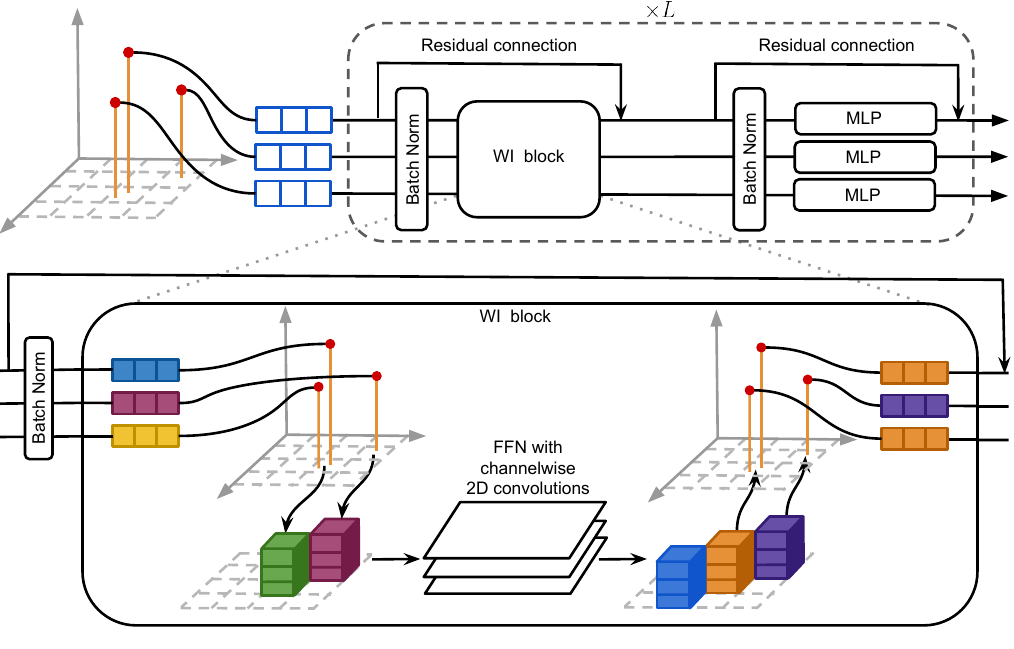}
\caption{\textbf{\ours backbone}. This 3D backbone takes as input point tokens, provided by an embedding layer (not represented), and updates these point representations $L$ times via a point token-mixing layer (containing the $\ourlayer$ block) followed by a channel-mixing layer. The $\ourlayer$ block consists of a 2D projection along one of the main axes, a feed-forward network (FFN) with two dense channel-wise 2D convolutions with a ReLU activation in the hidden layer, and a simple copy of the 2D features to the 3D points. The channel-mixing layer contains a batch-norm, a MLP shared accross each point, and a residual connection. The \ours backbone is free of any point downsampling or upsampling layer, farthest point sampling, nearest neighbor search, or sparse convolution.} 
\label{fig:method}
\end{center}
\end{figure*}

Lidar sensors deliver rich information about the 3D environment surrounding autonomous vehicles. Semantic segmentation of point clouds delivered by these lidars permits to autonomous vehicles to make sense of this 3D information in order to take proper and safe decisions. When studying the \href{http://www.semantic-kitti.org/tasks.html\#semseg}{leaderboard} of SemanticKITTI \cite{semkitti}, we rapidly notice that all the top methods leverage sparse 3D convolutions. For example, the recent work 2DPASS \cite{2dpass} relies on an adapted version of SPVCNN \cite{spvnas} which, once trained with the help of images of the scene captured synchronously with the lidar, is currently the state-of-the-art method. As another example, Cylinder3D \cite{Cylinder3d}, later improved in \cite{PVKD}, use sparse 3D convolutions on cylindrical voxels (particularly adapted to rotating lidars) with asymmetrical kernels suited to capture the geometry of the main objects in driving scenes.

Despite the undeniable success and efficiency of sparse convolutions, we seek here for 3D backbones which are free of them. Indeed, sparse convolutions remain available in a limited number of deep learning frameworks and hardware (essentially PyTorch and NVIDIA GPUs). One reason might be because they are challenging to implement efficiently \cite{torchsparse}. Another reason may be because they are not as widely used as, e.g., dense 2D convolutions, and are thus not the first to be implemented in a new framework. Therefore, we would like to construct a 3D backbone (i) built with tools more broadly available than sparse convolutions, but which (ii) can reach the level of performance of the top methods on automotive datasets, while (iii) remaining easy to implement and to use. This would offer a compelling alternative to sparse 3D backbone, especially when sparse convolutions are not available.

We actually construct a novel 3D backbone built almost exclusively with standard MLPs and dense 2D convolutions, both readily available in all deep learning frameworks thanks to their wide use in the whole field of computer vision. Our backbone architecture, \ours, is illustrated in \cref{fig:method}, and is inspired by the recent MLP-Mixer \cite{mlpmixer}. It takes as input a point cloud with a token associated to each point. All these point tokens are then updated by a sequence of layers, each containing a token-mixing step (made of dense 2D convolutions) and a channel-mixing step (made of a MLP shared across points).

In addition, we explain how to train \ours to make it reach the performance of the current best methods on automotive semantic segmentation benchmarks. The performance we obtain shows that standard MLPs and dense 2D convolutions, despite being a priori unfit for 3D segmentation, are sufficient to construct a 3D backbone reaching the state of the art.

Finally, \ours is at least as easy to implement and to tune as any other backbone. The implementation consists in repeated applications of basic layers directly on the point tokens (an example of complete implementation is available in the supplementary material). The performance increases with the network width and depth, until an eventual saturation.
The main hyperparameter to tune is the resolution of the 2D grid used for discretization before 2D convolution, but for which we observe stable results over a wide range of values (facilitating its tuning). 
The two most technical components to implement are reduced to: (i) the embedding layer used before \ours and providing the point tokens, and (ii) the 2D projections followed by feature discretizations (applied before dense 2D convolutions).

In summary, our contributions are the following.
\begin{itemize}[itemsep=-1pt,topsep=-1pt]
    \item We propose a novel and easy-to-implement 3D backbone for automotive point cloud semantic segmentation, which is essentially made of standard MLPs and dense 2D convolutions.
    \item We show that the hyperparameters of \ours are easy to tune: the performance increases with the width and depth, until a possible saturation; the performance is stable over a large range of 2D grid resolutions.
    \item We present how to train \ours to reach the performance of top-entries on two autonomous driving benchmarks: SemanticKITTI \cite{semkitti} and nuScenes \cite{nuscenes}. This shows that standard MLPs and dense 2D convolutions are actually sufficient to compete with the state of the art.
\end{itemize}

\section{Related Work}

We divide the related works into four categories: \emph{point-based methods}, that work directly on points and update point representations throughout the network; \emph{projection-based methods}, that project the points on a 2D grid at the input of the network, extract pixel-wise representations with a 2D network, and finally back-project the features in 3D for segmentation at the output of the network; \emph{sparse convolution-based methods}, which voxelize the point clouds and uses sparse convolutions; \emph{fusion-based methods}, which leverage different point cloud representations in parallel and fuse the corresponding features. 

\smallparagraph{Point-based methods.} PointNet \cite{pointnet} is the first method that appeared in this category, quickly followed by its improved version, PointNet++ \cite{pointnetpp}. Several methods then followed to improve the definition of point convolution, e.g., \cite{kpconv,dgcnn,fkaconv}, to scale to large point clouds by exploiting point clustering, e.g., \cite{SPGraph,PointNL}, to optimize point sampling, e.g., \cite{PointASNL, RandLANet}, or make point convolution faster to compute, e.g., \cite{latticenet}. Following the trend in image understanding, we also witness a growing amount of works, e.g., \cite{pointtransformer,fastpointtrans,strattrans}, exploiting transformer architectures, which are particularly suited to handle unordered set of points. Recently, {PointNext} \cite{pointnext} revisited and optimized PointNet++ with more modern tools and showed that it is still highly competitive in several benchmarks. In general, point-based methods are particularly effective to process dense point clouds such as those obtained with depth cameras in indoor scenes. These methods, unless combined with other point cloud representations, are seldomly used to process sparse outdoor lidar point clouds.

Among these point-based methods, let us discuss in more details two works which share some similarities with ours. The first work is PointMixer \cite{pointmixer} which takes inspiration from the MLP-Mixer \cite{mlpmixer}. Despite the same source of inspiration, we remark several fundamental differences with our work. (i)~The architecture differs significantly from \ours: PointMixer is a U-Net architecture with downsampling/upsampling layers, while we keep the resolution of point cloud fixed and do not use any skip connection between the early and last layers. (ii)~The spatial-mixing step is also fundamentally different as it is constructed using several sets of nearest neighbors points, while we use dense 2D convolutions. (iii)~The method is used on dense point clouds captured in indoor scenes. The second work is PointMLP \cite{pointmlp} which proposes a simple point-based network made only of MLPs. The PointMLP architecture is also very different from ours, starting with the spatial-mixing strategy which is done by aggregating information over sets of k-nearest neighbors. In addition, the application of PointMLP is limited to small scale point clouds for shape classification and part segmentation.

\smallparagraph{Projection-based methods.} Projection-based methods are more used to process point clouds acquired with rotating lidars than point-based approaches. By working almost entirely on 2D feature maps, they usually benefit from very fast computations. Yet, their performance remains below methods leveraging sparse convolutions. Among these methods, we find some using the spherical (range) projection \cite{rangenetpp} or the bird's eye view projection \cite{Polarnet}. Recent improvements have been achieved by making the convolution kernels better suited to the type of ``images'' produced by projection of the point clouds \cite{SqueezeSegV3}, by using techniques that reduce the loss information in the 2D encoder-decoder architectures \cite{salsaNext}, by solving an auxiliary tasks such as surface reconstruction \cite{SCSSnet}, adding a learned post-processing step in 3D \cite{KPRNet}, or exploiting vision transformers pretained on image datasets \cite{rangevit}.

\smallparagraph{Sparse convolution-based methods.} These type of methods leverage point cloud sparsity to reduce the computational and memory load. In particular, they  compute the result of the convolution only on occupied voxels \cite{minknet}. These methods become particularly efficient on autonomous driving scenes, e.g., when, adapting the shape of the voxels to the point sampling structure \cite{Cylinder3d}. Recently, some improvements have been obtained on these architectures by leveraging knowledge distillation techniques \cite{PVKD,sdseg3d}. Finally, some attention mechanisms are also now exploited on top of sparse convolution-based architectures to, e.g., adapt the classification layer to the input point cloud \cite{napl} or to improve feature quality \cite{AF2S3Net,gasn,svaseg,sphereformer}.

\smallparagraph{Fusion-based methods.} These methods try to combine the advantage of different point representations to improve semantic segmentation. They rely on, e.g., bird's eye view and range representations used in a sequence \cite{tornadonet}, or used in parallel for fusing deep features \cite{amvnet,gfnet}. Another strategy is to combine fine-grained features provided by point representation with high-level voxel representations \cite{spvnas,fusionnet,pcscnet}. RPVNet \cite{RPVNet} fuses features extracted at multiple layers of three different networks, each dealing with range, point or voxel representations.

%
\section{Our Method}
\label{sec:method}

\subsection{WaffleIron Backbone}

\smallparagraph{High-level description.} \ours is illustrated in \cref{fig:method}. It takes as input a point cloud with a $F$-dimensional token associated to each point. These point tokens, obtained by an embedding layer described in \cref{sec:practical_consideration}, are updated $L$ times thanks to token-mixing layers and channel-mixing layers. The core component of the token-mixing layer is our novel $\ourlayer$ block. It is made of a 2D projection along one of the main axes, a discretization of the features on a 2D grid, and a feed-forward network (FFN) with dense 2D convolutions. The channel mixing layer is essentially made of an MLP shared across each point. 

\smallparagraph{Formal definition.} \ours takes as input a point cloud with $N$ points whose Cartesian $xyz$-coordinates are denoted by $\vec{p}_i \in \mathbb{R}^3$, $i=1, \ldots, N$. Each point is associated with a point token $\vec{f}_i^{(0)} \in \mathbb{R}^{F}$ provided by a embedding layer (see \cref{sec:practical_consideration}). To simplify the following equations, we group all the point tokens in a large matrix $\ma{F}^{(0)}$ of size ${F \times N}$. These tokens are then transformed by a series of $L$ layers, each satisfying
\begin{alignat}{4}
\label{eq:spatial_mix}
& \ma{G}^{(\ell)} && = \ma{F}^{(\ell)} &&+ \ourlayer \, && ( \, {\rm BN} \, ( \, \ma{F}^{(\ell)} \, ) \, ), \\
\label{eq:channel_mix}
& \ma{F}^{(\ell+1)} && = \ma{G}^{(\ell)}  &&+ {\rm MLP} && ( \,{\rm BN} \, ( \, \ma{G}^{(\ell)} \, ) \, ), 
\end{alignat}
to obtain the deep point features $\ma{F}^{(L)} \in \mathbb{R}^{F \times N}$, then used to classify each point thanks to a single linear layer.\footnote{In our implementation, we also used two layerscale layers \cite{layerscale}: one after the $\ourlayer$ block and one after the MLP.} Eq.~\eqref{eq:spatial_mix} and Eq.~\eqref{eq:channel_mix} corresponds to the token-mixing step and channel-mixing step, respectively. ${\rm BN}$ denotes batch normalization. The ${\rm MLP}$ is applied point-wise and contains two layers with a ReLU activation after the first layer. 

The $\ourlayer$ block mixes the features spatially as illustrated in the lower part of \cref{fig:method}. It processes input 3D features $\ma{F} \in \mathbb{R}^{F \times N}$ in three steps to obtain the residual which satisfies $\ourlayer(\ma{F}) = {\rm Inflat} \circ {\rm Conv} \circ {\rm Flat} (\ma{F}).$ These three steps are described below.
\begin{enumerate}[itemsep=1pt,]
\item ${\rm Flat}(\cdot)$: Project (``flatten'') the points on one of the planes $(x, y)$, $(x, z)$ or $(y, z)$. Discretize the chosen plane into $M$ cells of size $\rho \times \rho$. Within each 2D cell, average the 3D features  of all points falling in this cell. We thus obtain the 2D feature map ${\rm Flat} (\ma{F}) \in \mathbb{R}^{F \times M}$.
\item ${\rm Conv}(\cdot)$: Process the 2D feature map ${\rm Flat} (\ma{F})$ with a feed-forward network (FFN) consisting of two layers of channel-wise 2D convolutions and a ReLU activation in the hidden layer. We obtain the 2D feature map ${\rm Conv} ( {\rm Flat} (\ma{F}))$.
\item ${\rm Inflat}(\cdot)$: For each 3D point, find the 2D cell into which this point falls into, and copy (``inflate'') the corresponding feature from ${\rm Conv} ( {\rm Flat} (\ma{F}))$. This yield the residual $\ourlayer(\ma{F}) \in \mathbb{R}^{F \times N}$.
\end{enumerate}

The name of our method, \ours, is inspired by the effect of the first step on the point cloud: it is flattened and imprinted with a regular 2D grid, as if it was compressed between the plates of a waffle iron.

\smallparagraph{${\rm \bf Flat}(\cdot)$ and ${\rm \bf Inflat}(\cdot)$ implementations.} The computations in ${\rm Flat}(\cdot)$ and ${\rm Inflat}(\cdot)$ are cheap. Both steps can be implemented using a sparse-dense matrix multiplication. It is sufficient to store a sparse matrix $\ma{S} \in \Rbb^{N \times M}$ with $N$ non-zero entries and structured as follows. For each 3D point $\vec{p}_i$: \emph{(a)} compute the index $j \in \{1, \ldots, M \}$ of the 2D cell into which this point falls into (by quantizing $\vec{p}_i$); \emph{(b)} set the entry in the $i^\th$ row and the $j^\th$ column of $\ma{S}$ to $1$. Then, the 2D feature map in the ${\rm Flat}(\cdot)$ step satisfies ${\rm Flat} (\ma{F}) = \ma{F} \, \ma{S} \oslash {\ma{N} \, \ma{S}}$, where $\ma{N} \in \Rbb^{F \times N}$ is a matrix where all entries are set to $1$ and $\oslash$ is the element-wise division. Note that $\ma{N} \, \ma{S}$ indicates the number of 3D points falling in each 2D cell, ensuring a proper average of 3D features falling in the same cells. Finally, the 3D residual $\ourlayer(\ma{F})$ obtained in the ${\rm Inflat}(\cdot)$ step satisfies $\ourlayer(\ma{F}) = {\rm Conv} ( {\rm Flat} (\ma{F})) \, \ma{S}^\adjoint.$

\subsection{Practical Considerations}
\label{sec:practical_consideration}

\smallparagraph{Choice of the projection plane.} In our proposed architecture, we repeatedly project along each main axis. Concretely, we sequentially project on planes $(x, y)$, $(x, z)$ and $(y, z)$ at layer $\ell=1$, $\ell=2$, and $\ell=3$, respectively, and repeat this sequence until layer $\ell = L$. In our experiments, we thus choose $L$ as a multiple of $3$. We nevertheless study the impact of different projection strategies in \cref{sec:projection}.

\smallparagraph{Resolution of the 2D grids.} For simplicity, we choose a single resolution $\rho \times \rho$ for all 2D grids used in the network. 

\smallparagraph{2D convolutions.} We use basic 2D kernels of size $3 \times 3$ for all layers throughout the network.

\smallparagraph{Embedding layer.} Let $\vec{h}_i$ denote the low-level features readily available at point $\vec{p}_i$, e.g., the height, range and lidar intensity of the point. Inspired by DGCNN \cite{dgcnn}, the embedding layer extracting the initial tokens $\vec{f}_i^{(0)}$ merges global and local information around each point:
\begin{align}
\label{eq:embedding}
\vec{f}_i^{(0)} = 
{\rm LN}\big( \, 
    [ {\rm LN} (\vec{h}_i) , \,
    \max_{j \in \set{N}_i} {\rm MLP} (\vec{h}_j - \vec{h}_i) ] \,
\big)
\end{align}
where ${\rm LN}$ denotes linear layers and $\set{N}_i$ the set of $k$ nearest points to $\vec{p}_i$. The features $\vec{h}_i$ are pre-normalized by a batch normalization layer before applying \eqref{eq:embedding}.

\subsection{Discussion}
\begin{table*}[t]
\small
\ra{1.2}
\newcommand*\rotext{\multicolumn{1}{R{65}{1em}}}
\setlength{\tabcolsep}{2.2pt}
\begin{center}
\begin{tabular}{l c c | c c c c c c c c c c c c c c c c c c c}
\toprule
    Method 
        & \rotext{SpConv free}\hspace{4mm}
        & \rotext{\bf mIoU\%}
        & \rotext{car}
        & \rotext{bicycle}
        & \rotext{motorcycle}
        & \rotext{truck}
        & \rotext{other-vehicle}
        & \rotext{person}
        & \rotext{bicyclist}
        & \rotext{motorcyclist}
        & \rotext{road}
        & \rotext{parking}
        & \rotext{sidewalk}
        & \rotext{other-ground}
        & \rotext{building}
        & \rotext{fence}
        & \rotext{vegetation}
        & \rotext{trunk}
        & \rotext{terrain}
        & \rotext{pole}
        & \rotext{traffic-sign}
\\ 
\midrule
    RandLA-Net~\cite{RandLANet} 
        & \cmark
        & 53.9
        & 94.2
        & 26.0 
        & 25.8 
        & 40.1
        & 38.9
        & 49.2 
        & 48.2 
        &  7.2 
        & 90.7
        & 60.3 
        & 73.7 
        & 20.4
        & 86.9 
        & 56.3 
        & 81.4 
        & 61.3 
        & 66.8
        & 49.2 
        & 47.7 
\\ 
    KPConv \cite{kpconv}
        & \cmark
        & 58.8
        & 96.0
        & 30.2
        & 42.5
        & 33.4 
        & 44.3
        & 61.5
        & 61.6
        & 11.8 
        & 88.8
        & 61.3
        & 72.7 
        & 31.6
        & 90.5
        & 64.2 
        & 84.8 
        & 69.2 
        & 69.1 
        & 56.4 
        & 47.4
\\
    SalsaNext \cite{salsaNext}
        & \cmark
        & 59.5
        & 91.9 
        & 48.3
        & 38.6 
        & 38.9 
        & 31.9 
        & 60.2
        & 59.0
        & 19.4
        & 91.7
        & 63.7 
        & 75.8
        & 29.1 
        & 90.2
        & 64.2
        & 81.8 
        & 63.6
        & 66.5 
        & 54.3 
        & 62.1 
\\
    NAPL \cite{napl}
        & \xmark
        & 61.6
        & 96.6
        & 32.3
        & 43.6
        & 47.3
        & 47.5
        & 51.1
        & 53.9
        & 36.5
        & 89.6
        & 67.1
        & 73.7
        & 31.2
        & 91.9
        & 67.4
        & 84.8
        & 69.8
        & 68.8
        & 59.1
        & 59.2
\\
    PCSCNet \cite{pcscnet}
        & \xmark
        & 62.7
        & 95.7 
        & 48.8
        & 46.2
        & 36.4
        & 40.6
        & 55.5
        & 68.4
        & \underline{55.9}
        & 89.1
        & 60.2
        & 72.4
        & 23.7
        & 89.3
        & 64.3
        & 84.2
        & 68.2
        & 68.1
        & 60.5
        & 63.9
\\
    KPRNet \cite{KPRNet}
        & \cmark
        & 63.1  
        & 95.5
        & 54.1
        & 47.9
        & 23.6
        & 42.6
        & 65.9
        & 65.0 
        & 16.5
        & 93.2
        & \textbf{73.9}
        & 80.6
        & 30.2
        & 91.7
        & 68.4
        & 85.7
        & 69.8
        & 71.2
        & 58.7
        & 64.1  
\\
    Lite-HDSeg \cite{Lite-HDSeg} 
        & \cmark
        & 63.8
        & 92.3
        & 40.0
        & 54.1
        & 37.7
        & 39.6
        & 59.2
        & 71.6
        & 54.1
        & 93.0
        & 68.2
        & 78.3
        & 29.3
        & 91.5
        & 65.0
        & 78.2
        & 65.8
        & 65.1
        & 59.5
        & 67.7
\\
    SVASeg \cite{svaseg}
        & \xmark
        & 65.2
        & 96.7
        & 56.4
        & 57.0
        & 49.1
        & 56.3
        & 70.6
        & 67.0
        & 15.4
        & 92.3
        & 65.9
        & 76.5
        & 23.6
        & 91.4
        & 66.1
        & 85.2
        & 72.9
        & 67.8
        & 63.9
        & 65.2
\\
    AMVNet \cite{amvnet}
        & \cmark
        & 65.3
        & 96.2 
        & 59.9 
        & 54.2 
        & 48.8 
        & 45.7 
        & 71.0 
        & 65.7 
        & 11.0 
        & 90.1 
        & 71.0
        & 75.8
        & 32.4
        & 92.4
        & 69.1
        & 85.6
        & 71.7
        & 69.6
        & 62.7
        & 67.2
\\
    GFNet \cite{gfnet}
        & \cmark
        & 65.4
        & 96.0 
        & 53.2 
        & 48.3 
        & 31.7 
        & 47.3 
        & 62.8 
        & 57.3 
        & 44.7 
        & \textbf{93.6}
        & \underline{72.5}
        & \textbf{80.8}
        & 31.2 
        & \textbf{94.0}
        & \textbf{73.9}
        & 85.2 
        & 71.1 
        & 69.3 
        & 61.8 
        & 68.0
\\
    JS3C-Net \cite{JS3CNet} 
        & \xmark
        & 66.0
        & 95.8
        & 59.3
        & 52.9
        & 54.3
        & 46.0
        & 69.5
        & 65.4
        & 39.9
        & 88.8
        & 61.9
        & 72.1
        & 31.9
        & 92.5
        & 70.8
        & 84.5
        & 69.8
        & 68.0
        & 60.7
        & 68.7
\\
    SPVNAS \cite{spvnas} 
        & \xmark
        & 66.4
        & 97.3
        & 51.5
        & 50.8
        & \textbf{59.8}
        & 58.8
        & 65.7
        & 65.2
        & 43.7
        & 90.2
        & 67.6
        & 75.2
        & 16.9
        & 91.3
        & 65.9
        & 86.1
        & 73.4
        & 71.0
        & 64.2
        & 66.9
\\
     2DPASS$^\star$ \cite{2dpass} 
        & \xmark
        & 67.4
        & 96.3
        & 51.1
        & 55.8
        & 54.9
        & 51.6
        & 76.8
        & 79.8
        & 30.3
        & 89.8
        & 62.1
        & 73.8
        & 33.5
        & 91.9
        & 68.7
        & 86.5
        & 72.3
        & 71.3
        & 63.7
        & 70.2
\\ 
    Cylinder3D \cite{Cylinder3d}  
        & \xmark
        & 67.8
        & 97.1 
        & 67.6
        & 64.0 
        & 50.8 
        & 58.6
        & 73.9 
        & 67.9 
        & 36.0 
        & 91.4 
        & 65.1 
        & 75.5 
        & 32.3 
        & 91.0 
        & 66.5 
        & 85.4 
        & 71.8 
        & 68.5 
        & 62.6 
        & 65.6
\\
    $($AF$)^2$-S3Net \cite{AF2S3Net} 
        & \xmark
        & 69.7
        & 94.5
        & 65.4
        & \textbf{86.8}
        & 39.2
        & 41.1
        & \textbf{80.7}
        & \underline{80.4}
        & \textbf{74.3}
        & 91.3
        & 68.8
        & 72.5
        & \textbf{53.5}
        & 87.9
        & 63.2
        & 70.2
        & 68.5
        & 53.7
        & 61.5
        & 71.0
\\
    RPVNet \cite{RPVNet} 
        & \xmark
        & 70.3
        & \textbf{97.6}
        & \underline{68.4}
        & 68.7
        & 44.2
        & \underline{61.1}
        & 75.9
        & 74.4
        & 43.4 
        & \underline{93.4}
        & 70.3
        & \underline{80.7}
        & 33.3
        & \underline{93.5}
        & 72.1
        & 86.5
        & \underline{75.1}
        & 71.7
        & 64.8
        & 61.4
\\
    SDSeg3D \cite{sdseg3d}
        & \xmark
        & 70.4
        & \underline{97.4}
        & 58.7
        & 54.2
        & 54.9
        & \textbf{65.2}
        & 70.2
        & 74.4
        & 52.2
        & 90.9
        & 69.4
        & 76.7
        & \underline{41.9}
        & 93.2
        & 71.1
        & 86.1
        & 74.3
        & 71.1
        & 65.4
        & 70.6
\\
    GASN \cite{gasn}
        & \xmark
        & 70.7
        & 96.9 
        & 65.8 
        & 58.0
        & \underline{59.3}
        & {61.0}
        & \underline{80.4}
        & \textbf{82.7}
        & 46.3
        & 89.8
        & 66.2
        & 74.6
        & 30.1
        & 92.3
        & 69.6
        & \textbf{87.3}
        & 73.0
        & \textbf{72.5}
        & \underline{66.1}
        & \underline{71.6}
\\
\rowcolor{gray!20}
     \bf \ours
        & \cmark
        & \underline{70.8}
        & 97.2
        & \textbf{70.0}
        & \underline{69.8}
        & 40.4
        & 59.6
        & 77.1
        & 75.5
        & 41.5
        & 90.6
        & 70.4
        & 76.4
        & 38.9
        & \underline{93.5}
        & \underline{72.3}
        & \underline{86.7}
        & \textbf{75.7}
        & 71.7
        & \textbf{66.2}
        & \textbf{71.9}
\\
    PVKD \cite{PVKD} 
        & \xmark
        & \textbf{71.2}
        & 97.0
        & 67.9
        & 69.3
        & 53.5
        & 60.2
        & 75.1
        & 73.5
        & 50.5
        & 91.8
        & 70.9
        & 77.5
        & 41.0
        & 92.4
        & 69.4
        & 86.5
        & 73.8
        & \underline{71.9}
        & 64.9
        & 65.8
\\
\bottomrule
\end{tabular}
\end{center}
\caption{Semantic segmentation performance on SemanticKITTI test set. The second column indicates if the method is free of sparse convolutions (SpConv). The best and second-best IoUs are bold and \underline{underlined}, respectively. The scores are obtained from the official \href{http://www.semantic-kitti.org/tasks.html\#semseg}{leaderboard} of SemanticKITTI when available, otherwise from the respective paper. Regarding 2DPASS$^\star$, we report the results of the baseline of \cite{2dpass} trained with lidar data but \emph{no} images, i.e., in the same setting as the other methods in this table. This table contains the score of methods published before the date of submission to ICCV23.}
\label{tab:semKITTI_test}
\end{table*}

\paragraph{Ease of implementation.} A PyTorch implementation of \ours is available in the supplementary material: it consists of repeated applications of basic layers directly on the point tokens, highlighting the implementation simplicity. We have successfully tested this implementation on NVIDIA GPUs but also, up to minor adaptations, on AMD GPUs, on which, as far as we know, no efficient implementation of sparse convolutions are readily available. This illustrates that \ours is easily usable on different hardwares.

We chose to keep the resolution of the point cloud constant all the way through the backbone. This avoids the implementation of point downsampling and upsampling layers, the tuning of the associated point sampling technique, and the multiple nearest neighbors searches that are usually involved. Despite the absence of such layers, \ours requires reasonable computing capacity: the model used to obtain our final result on SemanticKITTI can be trained on a single NVIDIA Tesla V100 GPU with 32 GB of memory. Nevertheless, improvements of \ours could include  downsampling layers to optimize the computation and memory loads.

Besides the embedding layer, the other most technical step to implement is the projection on  2D planes followed by feature discretization on a 2D grid. We greatly simplified this step: we project only along one of the main axes (so the projected coordinates are available without extra-computation); we use a single 2D grid resolution; feature discretization can be done by multiplication with a fixed (non-learnable) sparse matrix constructed thanks to a simple quantization of the point coordinates.

\paragraph{Ease of hyperparameter tuning.} We show in \cref{sec:hyperparameters} that the performance improves on all datasets when increasing the width $F$ and depth $L$ of \ours until a potential saturation. The final choice for these values could, for example, be guided essentially by the desired or available computation resources. The sole remaining parameter to tune in \ours is the resolution $\rho \times \rho$ of the 2D grid in the ${\rm Flat}(\cdot)$ step. The optimal value of this parameter is dataset-dependent but we noticed that results remain stable for a wide range of values, which makes intensive fine-tuning unnecessary. In particular, a resolution of $50 \ {\rm cm}$ is nearly optimal for both SemanticKITTI and nuScenes.

%
\section{Experiments}
\label{sec:experiments}

\subsection{Datasets}
We conduct experiments on two large-scale autonomous driving datasets: SemanticKITTI \cite{semkitti} and nuScenes \cite{nuscenes}.

\smallparagraph{SemanticKITTI.} This dataset contains 22 sequences where each point cloud is segmented into 19 semantic classes. We use the usual split where the first 11 sequences constitute the training set, except the 8$^\th$ sequence used for validation, and the last 11 sequences constitute the test set.

\smallparagraph{nuScenes.} Each point in this dataset \cite{nuscenes} is annotated with one of the 16 considered semantic classes. The dataset contains 1000 scenes acquired in Boston and Singapore. We use the official split with 700 scenes for training, 150 scenes for validation and 150 scenes for test.

\subsection{Implementation Details}
\begin{table*}[t]
\small
\ra{1.2}
\newcommand*\rotext{\multicolumn{1}{R{65}{1em}}}
\setlength{\tabcolsep}{3.5pt}
\begin{center}
\begin{tabular}{l c c | c c c c c c c c c c c c c c c c}
\toprule 
    Method
        & \rotext{SpConv free}\hspace{4mm}
        & \rotext{\bf mIoU\%}
        & \rotext{barrier}
        & \rotext{bicycle}
        & \rotext{bus}
        & \rotext{car}
        & \rotext{const. veh.}
        & \rotext{motorcycle}
        & \rotext{pedestrian}
        & \rotext{traffic cone}
        & \rotext{trailer}
        & \rotext{truck}
        & \rotext{driv. surf.}
        & \rotext{other flat}
        & \rotext{sidewalk}
        & \rotext{terrain}
        & \rotext{manmade}
        & \rotext{vegetation}
\\
\midrule
    $($AF$)^2$-S3Net \cite{AF2S3Net}
        & \xmark
        & 62.2
        & 60.3 
        & 12.6 
        & 82.3 
        & 80.0 
        & 20.1 
        & 62.0 
        & 59.0 
        & 49.0 
        & 42.2 
        & 67.4 
        & 94.2 
        & 68.0 
        & 64.1 
        & 68.6 
        & 82.9 
        & 82.4
\\
    RangeNet++ \cite{rangenetpp}
        & \cmark
        & 65.5
        & 66.0 
        & 21.3 
        & 77.2 
        & 80.9 
        & 30.2 
        & 66.8 
        & 69.6 
        & 52.1 
        & 54.2 
        & 72.3 
        & 94.1 
        & 66.6 
        & 63.5 
        & 70.1 
        & 83.1 
        & 79.8
\\
    PolarNet \cite{Polarnet}
        & \cmark
        & 71.0
        & 74.7 
        & 28.2 
        & 85.3 
        & 90.9 
        & 35.1 
        & 77.5 
        & 71.3 
        & 58.8 
        & 57.4 
        & 76.1 
        & 96.5 
        & 71.1 
        & 74.7 
        & 74.0 
        & 87.3 
        & 85.7
\\
    SalsaNext \cite{salsaNext}
        & \cmark
        & 72.2
        & 74.8 
        & 34.1 
        & 85.9 
        & 88.4 
        & 42.2 
        & 72.4 
        & 72.2 
        & 63.1 
        & 61.3 
        & 76.5 
        & 96.0 
        & 70.8 
        & 71.2 
        & 71.5 
        & 86.7 
        & 84.4
\\
    SVASeg \cite{svaseg}
        & \xmark
        & 74.7
        & 73.1
        & 44.5
        & 88.4
        & 86.6
        & 48.2
        & 80.5
        & 77.7
        & 65.6
        & 57.5
        & 82.1
        & 96.5
        & 70.5
        & 74.7
        & 74.6
        & 87.3
        & 86.9
\\
    AMVNet \cite{amvnet}
        & \cmark
        & 76.1
        & \underline{79.8}
        & 32.4 
        & 82.2 
        & 86.4 
        & \textbf{62.5}
        & 81.9 
        & 75.3 
        & \textbf{72.3}
        & \textbf{83.5}
        & 65.1 
        & \underline{97.4}
        & 67.0 
        & \underline{78.8}
        & 74.6
        & \underline{90.8}
        & \underline{87.9}
\\
    GFNet \cite{gfnet}
        & \cmark
        & 76.1
        & \textbf{81.1}
        & 31.6 
        & 76.0 
        & 90.5 
        & \underline{60.2}
        & 80.7 
        & 75.3 
        & \underline{71.8}
        & \underline{82.5}
        & 65.1 
        & \textbf{97.8}
        & 67.0 
        & \textbf{80.4}
        & \textbf{76.2}
        & \textbf{91.8}
        & \textbf{88.9}
\\
    Cylinder3D \cite{Cylinder3d}
        & \xmark
        & 76.1
        & 76.4 
        & 40.3 
        & 91.2 
        & \textbf{93.8}
        & 51.3
        & 78.0 
        & 78.9 
        & 64.9 
        & 62.1 
        & 84.4
        & 96.8 
        & 71.6
        & 76.4
        & 75.4
        & 90.5 
        & 87.4
\\
    2DPASS$^{\star\dagger}$ \cite{2dpass}
        & \xmark
        & 76.2
        & 75.3
        & 43.5
        & \textbf{95.3}
        & 91.2
        & 54.5
        & 78.9
        & 72.8
        & 62.1
        & 70.0
        & 83.2
        & 96.3
        & 73.2
        & 74.2
        & 74.9
        & 88.1
        & 85.9
\\
    RPVNet \cite{RPVNet}
        & \xmark
        & 77.6
        & 78.2
        & 43.4
        & 92.7
        & \underline{93.2}
        & 49.0
        & 85.7
        & 80.5
        & 66.0
        & 66.9
        & 84.0
        & 96.9
        & 73.5
        & 75.9
        & \underline{76.0}
        & 90.6
        & \textbf{88.9}
\\
\rowcolor{gray!20}
    \bf\ours (ours)
        & \cmark
        & 77.6
        & 78.7
        & 51.3
        & 93.6 
        & 88.2
        & 47.2
        & 86.5
        & 81.7
        & 68.9
        & 69.3
        & 83.1
        & 96.9
        & 74.3
        & 75.6 
        & 74.2
        & 87.2 
        & 85.2
\\
    SDSeg3D \cite{sdseg3d}
        & \xmark
        & 77.7
        & 77.5 
        & 49.4 
        & 93.9 
        & 92.5 
        & 54.9
        & 86.7 
        & 80.1 
        & 67.8 
        & 65.7 
        & \underline{86.0}
        & 96.4 
        & 74.0
        & 74.9
        & 74.5
        & 86.0
        & 82.8
\\
    SDSeg3D$^\dagger$ \cite{sdseg3d}
        & \xmark
        & \underline{78.7}
        & 78.2
        & \underline{52.8}
        & \underline{94.5}
        & 93.1
        & 54.5
        & \underline{88.1}
        & \underline{82.2}
        & 69.4
        & 67.3
        & \textbf{86.6}
        & 96.4
        & \underline{74.5}
        & 75.2
        & 75.3
        & 87.1
        & 84.1
\\
\rowcolor{gray!20}
    \bf WaffleIron$^\dagger$ (ours)
        & \cmark
        & \textbf{79.1} 
        & \underline{79.8}
        & \textbf{53.8}
        & 94.3 
        & 87.6 
        & 49.6 
        & \textbf{89.1} 
        & \textbf{83.8} 
        & 70.6 
        & 72.7 
        & 84.9
        & 97.1
        & \textbf{75.8}
        & 76.5 
        & 75.9 
        & 87.8 
        & 86.3 
\\
\bottomrule
\end{tabular}
\end{center}
\caption{Semantic segmentation performance on nuScenes validation set. The second column indicates if the method is free of sparse convolutions (SpConv). Best and second-best scores are bold and underlined. The scores of each method are obtained from their respective paper, except for RangeNet++, PolarNet, SalsaNext for which they were obtained from \cite{Cylinder3d}, and for AMVNet obtained from \cite{RPVNet}. Regarding 2DPASS$^{\star}$, we report the scores obtained for the network trained using lidar data and \emph{no images}, i.e., in the same setting as the other methods in this table. Test time augmentations (TTA), indicated by $^\dagger$, are used in some methods; we thus report the score of \ours with and without TTA. This table contains the score of methods published before the date of submission to ICCV23.}
\label{tab:nuscenes_val_set}
\end{table*}

During training and test, the point clouds are slightly downsampled by keeping only one point per voxel of size 10 cm. We use mixed precision for computations. At test time, the predicted labels are propagated to all points of the original point cloud by nearest neighbor interpolation.

\smallparagraph{Training.} To control the memory usage and facilitate batch processing, we pre-process the point cloud as follows. We keep the size $M$ of the 2D grids used in the $\ourlayer$ blocks fixed. This is achieved by cropping the input point cloud to a fixed range. On SemanticKITTI, we use a range of $(-50 \, {\rm m}, \, 50 \, {\rm m})$ along the $x,y$ axes and $(-3 \, {\rm m}, \, 2 \, {\rm m})$ along the $z$-axis, as in \cite{Cylinder3d}. On nuScenes, we use the same range of along the $x,y$ axes and $(-5 \, {\rm m}, \, 5 \, {\rm m})$ along the $z$-axis. We also keep the number of points $N=20\ 000$ fixed. If the input point cloud has a size larger than $N$, then we pick a point at random and keep its closest $N-1$ points, otherwise the point cloud is zero padded.

All models are  trained using AdamW for $45$ epochs, with a weight decay of $0.003$, a batch size of $4$, and a learning rate scheduler with a linear warmup phase from $0$ to $0.001$ during the first $4$ epochs followed by a cosine annealing phase that decreases the learning rate to $10^{-5}$ at the end of the last epoch. The loss is the sum of the cross-entropy and the Lov\'asz loss \cite{lovasz}. The point tokens are computed with $16$ nearest neighbors in the embedding layer \eqref{eq:embedding}. We apply classical point cloud augmentations on nuScenes and SemanticKITTI: random rotation around the $z$-axis, random flip of the direction of the x and y-axis, and random scaling. Unless mentioned otherwise, we also use stochastic depth~\cite{stocdepth} with a layer drop probability of 0.2.

\smallparagraph{Test and validation.} Because the range along each axis considered at train time is sufficiently large to contain nearly the whole point clouds, we continue cropping the points clouds on the same range during validation and test. The labels of the points outside the range are obtained by nearest neighbors interpolation. We use all the input points after voxel downsampling (hence do not constraint $N$) during test and validation. Some methods leverage test time augmentations, e.g., \cite{Cylinder3d,PVKD,2dpass}; when applied, we average the softmax pointwise probabilities obtained with 10 different augmentations (random rotation, flip and stochastic depth activated). We do \emph{not} use model ensemble to boost the test or validation performance.

\smallparagraph{Input features.} Unless mentioned otherwise, the input feature $\vec{h}_i$ to the embedding layer is a $5$-dimensional vector which contains the lidar intensity, the Cartesian coordinates $xyz$ and range of the corresponding point $\vec{p}_i$.

\subsection{Performance on Autonomous Driving Datasets}
\label{sec:bench_ad}

On both datasets, we train a \ours backbone with $L=48$ layers. We use $F=256$-dimensional point tokens and a grid resolution $\rho$ of $40$ cm  on SemanticKITTI. We use $F=384$ and $\rho = 60$ cm on nuScenes. These choices of hyperparameters are justified in the next sections.

\smallparagraph{SemanticKITTI.} We evaluate our method on the test split. We adopt the training and inference practices used by the best performing techniques, e.g., \cite{RPVNet,Cylinder3d,PVKD,2dpass}. In particular, the model is trained using both the training and validation splits, and test time augmentations are used at inference. In addition to the classical rotation, flip, scaling augmentations during training, \cite{RPVNet,2dpass} use instance cutmix augmentations. Taking also inspiration from the suggestions made in the official code repository of \cite{PVKD}, we combine instance cutmix with polarmix \cite{polarmix}. We provide further details about instance cutmix and polarmix in \cref{sec:bonus}.

We present the results obtained on the test set in \cref{tab:semKITTI_test}. \ours is ranked second in term of global mIoU, just $0.4$ point away from PVKD. We surpasses the mIoU obtained with popular methods such as Cylinder3D and SPVNAS. It is interesting to notice that \ours is among the best methods in the segmentation of small and rare objects such as bicycles, motorcycles, poles and traffic-signs. The take-home message is that \ours is among the top performing methods on SemanticKITTI, making it a compelling alternative if, e.g., one is constrained to using regular deep network layers.

\smallparagraph{nuScenes.} The model is trained on the official training split. We present in \cref{tab:nuscenes_val_set} the scores obtained by \ours and other methods on the validation set.
Once again the results show that \ours can reach the current best mIoUs. As before, it is interesting to notice that \ours performs well on rare and small objects such as bicyles, motorcycles and pedestrians. It confirms it is possible to reach the top of the leaderboard on nuScenes with \ours.

\subsection{Sensitivity to Hyperparameters}
\label{sec:hyperparameters}

\begin{figure}[t]
\centering
\begin{tikzpicture}
    \tikzstyle{every node}=[font=\scriptsize]
    \begin{axis}[
        width=0.93\linewidth,
        font=\scriptsize,
        xtick={20,30,40,50,60,70,80},
        xmin=20, xmax=80,
        ytick={50,55,60,65,70,75,80},
        ymin=50, ymax=80,
        ylabel=mIoU\%,
        xlabel=$\rho \ ({\rm cm})$,
        label style={font=\small},
        tick label style={font=\scriptsize},
        grid=major,
    ]
    \addplot[densely dotted, color=teal, mark=x, line width=0.5mm] 
    coordinates 
    {
    (20,61.7)
    (30,63.3)
    (40,64.3)
    (60,65.2)
    (80,64.6)
    };
    \label{p1}
    \addplot[color=teal, mark=x, line width=0.5mm] 
    coordinates 
    {
    (20,73.5)
    (30,75.0)
    (40,74.9)
    (60,75.2)
    (80,74.9)
    };
    \label{p2}
    \addplot[densely dotted, color=Orange, mark=+, line width=0.5mm] 
    coordinates 
    {
    (20,57.4)
    (30,57.7)
    (40,58.2)
    (60,57.9)
    (80,56.8)
    };
    \label{p3}
    \addplot[color=Orange, mark=+, line width=0.5mm] 
    coordinates 
    {
    (20,62.5)
    (30,62.5)
    (40,62.6)
    (60,61.6)
    (80,60.2)
    };
    \label{p4}
    \end{axis}
    \node [draw,fill=white] at (rel axis cs: 0.33,0.92) {\shortstack[l]
    {
        \ref{p1} nuScenes - WaffleIron-6-64 \\
        \ref{p2} nuScenes - WaffleIron-12-256
    }
    };
    \node [draw,fill=white] at (rel axis cs: 0.31,0.08) {\shortstack[l]
    {
        \ref{p3} KITTI - WaffleIron-6-64 \\
        \ref{p4} KITTI - WaffleIron-12-256
    }
    };
\end{tikzpicture}
\caption{Influence of the grid resolution $\rho$ on the performance of \ours. We train each backbone on the training set of nuScenes or SemanticKITTI and compute the mIoU on the corresponding validation set. We report the average mIoU\% obtained at the last training epoch of two independent runs.}
\label{fig:sensitivity_grid}
\end{figure}
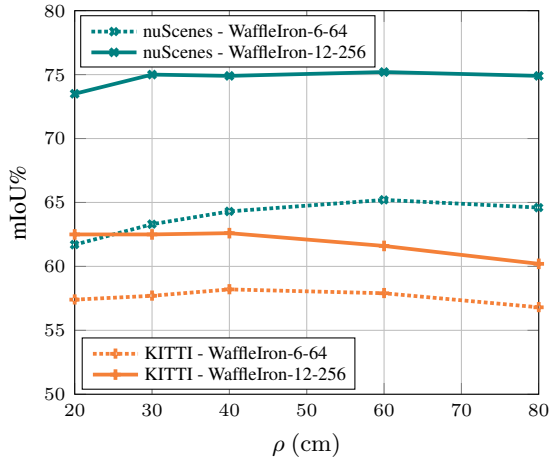

We denote by \ours-$L$-$F$ a backbone with $L$ layers and $F$-dimensional point tokens. We only use here $3$-dimensional vectors $\vec{h}_i$ (lidar intensity, height and range of $\vec{p}_i$) and do not use stochastic depth for training. We justify the use of $5$-dimensional vectors $\vec{h}_i$ and stochastic depth in the next section.

\smallparagraph{2D grid resolution.} We study the impact of $\rho$ on each dataset for two versions of our network: \ours-6-64  and \ours-12-256. We notice in \cref{fig:sensitivity_grid} that the performance are stable for a large range of grid resolutions. On nuScenes, the mIoU\% varies by at most one point for $\rho$ between $40 \ {\rm cm}$ and $80 \ {\rm cm}$ with a maximum reached at $60 \ {\rm cm}$. Similarly, on SemanticKITTI, the mIoU\% varies by at most one point for $\rho$ between $20 \ {\rm cm}$ and $60 \ {\rm cm}$, with a maximum reached at $40 \ {\rm cm}$. In summary, \ours is only mildly sensitive to the grid resolution, and, therefore, can accommodate a coarse tuning of this parameter. In particular, $\rho = 50 \ {\rm cm}$ could be a good default value to accommodate nearly optimally both datasets.

\begin{table}[t]
\begin{center}
\ra{1.1}
\setlength{\tabcolsep}{8pt}
\begin{tabular}{@{}l c c c c@{}}
\toprule
    & \multicolumn{4}{c}{nuScenes ($\rho = 60 {\rm cm}$)}
\\
\cmidrule(lr){2-5}
    & $L=6$
    & $L=12$ 
    & $L=24$ 
    & $L=48$
    \\
\midrule
$F=64$
    & \setstretch{0.7}\makecell{65.2}
    & -
    & -
    & -
    \\
$F=128$
    & \setstretch{0.7}\makecell{70.8}
    & -
    & -
    & -
    \\
$F=256$
    & \setstretch{0.7}\makecell{73.2}
    & \setstretch{0.7}\makecell{75.2}
    & \setstretch{0.7}\makecell{75.4}
    & \setstretch{0.7}\makecell{76.1}
\\
\midrule
& \multicolumn{4}{c}{KITTI ($\rho = 40 {\rm cm}$)}
\\
\cmidrule(lr){2-5}
    & $L=6$ 
    & $L=12$ 
    & $L=24$ 
    & $L=48$
\\
\midrule
$F=64$
    & \setstretch{0.7}\makecell{58.2}
    & -
    & -
    & -
\\
$F=128$
    & \setstretch{0.7}\makecell{61.4}
    & -
    & -
    & -
\\
$F=256$
    & \setstretch{0.7}\makecell{61.8}
    & \setstretch{0.7}\makecell{62.6}
    & -
    & \setstretch{0.7}\makecell{62.5}
\\
\bottomrule
\end{tabular}
\end{center}
\caption{Influence of the width $F$ and depth $L$ on the performance of \ours. We train each backbone on the training set of nuScenes or SemanticKITTI and compute the mIoU on the corresponding validation set. We report the average mIoU\% obtained at the last training epoch of two independent runs.}
\label{tab:perf_vs_L_F}
\end{table}

\smallparagraph{Choice of $F$ and $L$.} We study in \cref{tab:perf_vs_L_F} the impact of increasing $L$ and $F$ in \ours. We notice the same behavior on both datasets with an increase of performance as both $L$ and $F$ increases, with the start of a saturation on SemanticKITTI. On SemanticKITTI, we did not notice any improvement or degradation for $F \geq 256$ at $L=48$. We chose \ours-48-256 to obtain our result on the test set. On nuScenes, we were able to improve the results when using $F=384$ (see supp.~mat.), hence our choice in \cref{sec:bench_ad}.

\subsection{Regularizations and input features}
\label{sec:bonus}
\pgfplotstableread[row sep=\\,col sep=&]{
    interval			& Bas. & Bas.Aug & xyz   & depth \\
    {car} 			& 94.8 & 95.8      & 95.7 & 95.9 \\
    {bicycle}		& 44.2 & 57.9      & 59.8 & 59.6 \\
    {motorcycle}		& 65.9 & 76.1      & 81.1 & 81.6 \\	
    {truck}			& 79.4 & 82.4      & 76.6 & 73.5 \\
    {other-vehicle}	& 33.4 & 52.7      & 51.8 & 57.2 \\
    {person}		& 63.8 & 78.2      & 80.6 & 81.0 \\
    {bicyclist}		& 83.3 & 93.0      & 92.6 & 92.3 \\
    {road}			& 95.0 & 95.3      & 95.5 & 95.6 \\
    {parking}		& 47.5 & 50.7      & 51.7 & 51.0 \\	
    {sidewalk}		& 83.2 & 83.5      & 83.9 & 83.7 \\	
    {other-ground}	&  0.0  & 3.2	  &  5.1  &  5.6 \\
    {building}		& 90.5 & 91.4      & 92.1 & 92.2 \\
    {fence}			& 61.0 & 64.6      & 68.0 & 67.8 \\
    {vegetation}		& 88.7 & 87.6      & 88.4 & 88.1 \\
    {trunk}			& 68.1 & 69.4      & 71.6 & 73.5 \\
    {terrain}		& 76.9 & 73.5      & 74.9 & 73.7 \\	
    {pole}			& 63.4 & 63.9      & 64.4 & 65.9 \\
    {traffic-sign}		& 48.7 & 50.2      & 51.3 & 52.5 \\
    mIoU			& 62.5 & 66.8      & 67.6 & 68.0 \\
    }\mydata

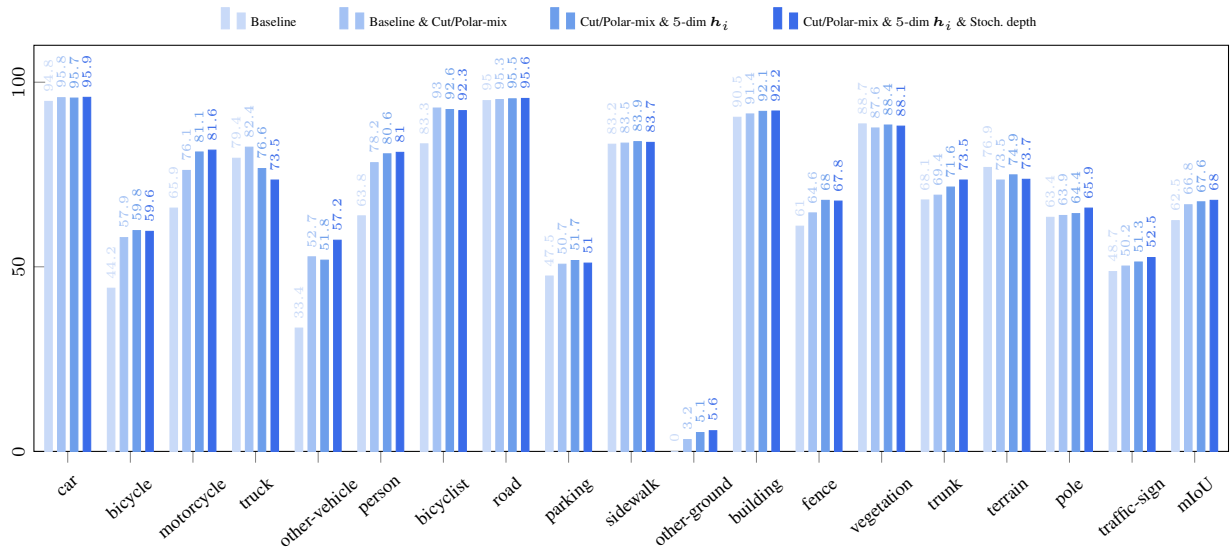
\begin{figure*}[t]
\centering
\begin{tikzpicture}
    \begin{axis}[
            ybar,
            bar width=.1cm,
            width=1.0\textwidth,
            height=.4\textwidth,
            font=\tiny,
            tick label style={font=\scriptsize},
            x tick label style={font=\scriptsize, rotate=40},
            y tick label style={font=\scriptsize, rotate=90},
            legend style={at={(0.5,1.01)}, anchor=south,legend columns=-1,draw=none,/tikz/every even column/.append style={column sep=0.5cm}},
            symbolic x coords={car,bicycle,motorcycle,truck,other-vehicle,person,bicyclist,road,parking,sidewalk,other-ground,building,fence,vegetation,trunk,terrain,pole,traffic-sign,mIoU},
            xtick=data,
            ytick={0,50,100},
            nodes near coords,
            nodes near coords align={vertical},
            every node near coord/.append style={rotate=90, anchor=west},
            ymin=0,ymax=100,
            enlarge x limits={0.03},
            enlarge y limits={upper,value=0.1},
            xtick pos=left,
            ytick pos=left,
        ]
        \addplot[color=blue4,fill=blue4] table[x=interval,y=Bas.]{\mydata};
        \addplot[color=blue3,fill=blue3] table[x=interval,y=Bas.Aug]{\mydata};
        \addplot[color=blue2,fill=blue2] table[x=interval,y=xyz]{\mydata};
        \addplot[color=blue1,fill=blue1] table[x=interval,y=depth]{\mydata};
        \legend{
            Baseline,
            Baseline \& Cut/Polar-mix,
            Cut/Polar-mix \& $5$-dim $\vec{h}_i$,
            Cut/Polar-mix \& $5$-dim $\vec{h}_i$ \& Stoch.~depth,
        }
    \end{axis}
\end{tikzpicture}
\vspace*{-2mm}
\caption{Influence of polarmix, instance cutmix, input vectors $\vec{h}_i$, and stochastic depth on the performance of \ours on SemanticKITTI. We train and evaluate \ours-48-256 backbones on the official train and validation set, respectively. We report the average mIoU\% obtained at the last training epoch of two independent runs. To improve readability, we omitted the IoU\% on motorcyclist, which varies between $0.0$ and $1.3$.}
\label{fig:augment}
\end{figure*}

In this section, we show the benefit of using more regularizations via data augmentations with instance cutmix and polarmix (only on SemanticKITTI), and via the use of stochastic depth during training. We also justify the use of $5$-dimensional input vectors $\vec{h}_i$, as opposed to the 3-dimensional $\vec{h}_i$ used in \cref{sec:hyperparameters}. We present here the results on SemanticKITTI. A similar study is available in the supp.~mat.~for nuScenes. ``Baseline'' refers to a \ours-48-256 backbone trained with $3$-dimensional input vectors (lidar intensity, height and range), no stochastic depth, no instance cutmix or polarmix.

\smallparagraph{Instance cutmix \& polarmix on SemanticKITTI.} Following \cite{RPVNet,2dpass}, we use instance cutmix on rare-class objects to improve the segmentation performance on SemanticKITTI. In our implementation, we extract all instances of the following classes: bicycle, motorcycle, person, bicyclist, other vehicles. During training, we randomly select at most $40$ instances of each class; we apply a random rotation around the $z$-axis, a random flip along the direction of the $x$ or $y$-axes, and a random rescaling on each instance; we place each instance at a random location on a road, parking or sidewalk. We did not apply instance cutmix on motorcyclists. Indeed, our method (like many others) reaches very low score on this class on the validation set. Tuning instance cutmix on motorcyclists is thus impossible as we cannot measure its beneficial or adverse effect. We therefore make the choice to not apply instance cutmix on motorcyclists. In addition, we also use polarmix \cite{polarmix} on the same classes as instance cutmix.

The impact of these augmentations is presented in \cref{fig:augment}. The mIoU\% improves from 62.5 to 66.8, with, as expected, most of the improvement due to a large boost of performance in the classes used for these augmentations.

\smallparagraph{Input features \& stochastic depth.} We start by showing the interest of using $5$-dimensional input vectors $\vec{h}_i$ (intensity, $x$, $y$, $z$, and range of $\vec{p}_i$) instead of $3$-dimensional input vectors (intensity, height=$z$, range). The impact of this change of input vector is presented in \cref{fig:augment} on top instance cutmix and polarmix: the mIoU\% increases from $66.8$ to $67.6$ with an improvement on most classes. Finally, using stochastic depth on top of all presented recipes permits us to achieve our best mIoU\% of $68.0$ on the validation set.

\subsection{Inference time}

We report the \textit{inference} time (embedding + WaffleIron + classification) of WaffleIron-48-256 on nuScenes and SemanticKITTI in \cref{tab:new_time}. Note that, here, we used the function {\ttfamily\small torch.gather} instead of a matrix-vector multiplication with sparse matrices to implement ${\rm Inflat}(\cdot)$ and the batch normalization in Eq.\,(2) were merged with the first linear layer of the following MLP.

The inference time of WaffleIron-48-256 is comparable to other sparse convolution-based methods on nuScenes and a bit slower ($\times$ 1.7) than the well-known MinkUNet34 on SemanticKITTI. Note that the modified SPVCNN in 2DPASS (SPVCNN$^\dagger$) is wider and deeper on nuScenes than on SemanticKITTI, hence the faster running time on the latter.

\begin{table}[t]
\begin{center}
\footnotesize
\begin{tabular}{@{}lccccc@{}}
\toprule
Time (ms)
    & Mink34
    & Mink18
    & {\setstretch{0.7}{\makecell{SPVCNN\\(orig.)}}}
    & {\setstretch{0.7}{\makecell{SPVCNN$^\dagger$\\(2DPASS)}}}
    & Ours
    \\
\midrule
nuScenes
    & \hphantom{0}94 
    & 66
    & \hphantom{0}74
    & 94
    & \hphantom{0}92
    \\
SemKITTI 
    & 114
    & 91
    & 104
    & 80
    & 193
    \\
\bottomrule
\end{tabular}
\end{center}
\caption{Inference time of several backbones and WaffleIron-48-256 (ours: embedding + backbone + classification) estimated on the validation sets of nuScenes and semanticKITTI, using a batch size of $1$ and a NVIDIA GeForce RTX 2080 Ti.}
\label{tab:new_time}
\end{table}
%

\subsection{Other choices of projection strategy}
\label{sec:projection}
\begin{table}[t]
\begin{center}
\ra{1.2}
\setlength{\tabcolsep}{3pt}
\begin{tabular}{@{}l c c c c c c c c c c c c@{}}
\toprule
    & \multicolumn{4}{c}{nuScenes ($\rho = 60 {\rm cm}$)}
\\
\cmidrule(lr){2-5}
\multicolumn{1}{c}{Projection}
    & Baseline 
        & Reverse
        & Parallel
        & BEV
    \\
\midrule
\ours-12-256
    & 75.2			
    & 75.0
    & 73.4
    & 74.8
    \\
\ours-48-256
    &   76.1
    & -
    & -
    & 76.0
\\
\midrule
    & \multicolumn{4}{c}{KITTI ($\rho = 40 {\rm cm}$)}
\\
\cmidrule(lr){2-5}
\multicolumn{1}{c}{Projection}
    & Baseline 
    & Reverse
    & Parallel
    & BEV
\\
\midrule
\ours-12-256
    & \setstretch{0.7}{\makecell{62.6}}
    & \setstretch{0.7}{\makecell{60.9}}
    & \setstretch{0.7}{\makecell{61.2}}
    & \setstretch{0.7}{\makecell{63.3}}
\\
\ours-48-256
    & \setstretch{0.7}{\makecell{62.5}}
    & -
    & -
    & \setstretch{0.7}{\makecell{63.7}}
\\
\ours-48-256$^\dagger$
    & \setstretch{0.7}{\makecell{66.8}}
    & -
    & -
    & \setstretch{0.7}{\makecell{66.8}}
\\
\bottomrule
\end{tabular}
\end{center}
\caption{Influence of the projection strategy on the performance of \ours. We train each backbone on the training set of nuScenes or SemanticKITTI and compute the mIoU on the corresponding validation set. We report the average mIoU\% obtained at the last training epoch of two independent runs. $^\dagger$ indicates that the backbone was trained with instance cutmix and polarmix augmentations.}
\label{tab:projection}
\end{table}

We present in \cref{tab:projection} the effect of using different projection strategies in our $\ourlayer$ block. These strategies are the following. \emph{Baseline} corresponds to the sequence of projections described in \cref{sec:practical_consideration}, i.e., used to produce all our results so far. \emph{Reverse} consists in reversing the order of the projections used in \emph{Baseline}. \emph{Parallel} consists in performing three projections on $(x, y)$, $(x, z)$ and $(y, z)$ in parallel at each layer. The projected feature maps are processed by different 2D FFNs. The resulting feature maps are then inflated, added together, and used as residual in \eqref{eq:spatial_mix}. We choose to compare this projection strategy to the others while keeping the number of 2D convolutions fixed. The actual depth of the network with this strategy is thus divided by three. \emph{BEV} consists in projecting on the $(x, y)$ plane at all layers. All experiments are conducted in the same setting as in \cref{sec:hyperparameters}.

First, reversing the sequence of projections has almost no effect on nuScenes where the mIoU\% decrease by 0.2 point. We notice however a decrease in mIoU on SemanticKITTI. We will see below that, on this dataset, projecting only on $(x, y)$ permits to improve the performance in absence of strong augmentations. We suppose that starting by projecting on $(x, y)$ has a positive effect thanks to, maybe, a better start in identifying the main structures.

Second, computing multiple projections in parallel is less optimal than computing them in a series: we loose -1.4 point and -1.8 point in mIou\% with respect to the baseline on SemanticKITTI and nuScenes, respectively.

Finally, projecting only in BEV has a negligible impact on the average mIoU on nuScenes: we loose at most -0.4 point in mIou\% with respect to the baseline sequence of projections. We explain this result because most structures and objects remain well identifiable in the bird's eye view in autonomous driving datasets. On SemanticKITTI, the baseline sequence of projections yields the same performance than the BEV projections only if strong data augmentations (instance cutmix and polarmix) are used during training. In absence of these augmentations, projecting only in bird's eye view might have played the role of a regularization which helped the generalization to unseen data.

%
\section{Conclusion}
\label{sec:conclusion}

We proposed \ours, a novel and easy-to-implement 3D backbone for automotive point cloud semantic segmentation, which is essentially made of standard MLPs and dense 2D convolutions. We showed that its hyperparameters are easy to tune and that it can reach the mIoU of top entries on two autonomous driving benchmarks.

Thanks to the use of dense 2D convolutions, we foresee other potential applications where \ours could be useful. In particular, the tasks semantic completion and or occupancy completion, see, e.g., \cite{SCSSnet,lmscnet}, where the $\ourlayer$ layer could be used to densify the input point cloud.

\smallparagraph{Acknowledgments.} We thank the Astra-vision team at Inria Paris for helpful discussions and insightful comments.  We also acknowledge the support of the French Agence Nationale de la Recherche (ANR), under grant ANR-21-CE23-0032 (project MultiTrans). This work was granted access to the HPC resources of CINES under the allocation GDA2213 for the Grand Challenges AdAstra GPU made by GENCI.

%
{\small
\bibliographystyle{ieee_fullname}
\bibliography{egbib}
}

%
\appendix
\section{\ours Implementation}

We present in \cref{code:waffleiron} an example of a code implementing the WaffleIron backbone in PyTorch \cite{pytorch}. We recall that this backbone takes as input point tokens provided by an embedding layer and outputs updated point tokens used in a linear classification layer for semantic segmentation. The implementation consists of applications of basic layers directly to each point tokens (batch normalizations, 1D and 2D convolutions, matrix-vector multiplications).

The step which is, maybe, the most technical to implement is the construction of the sparse matrices (line 56 of \cref{code:waffleiron}) for projections from 3D to 2D. For completeness, we provide the corresponding code as well in \cref{code:spmat}. Creating these sparse matrices requires computing the mapping between each 3D point and each 2D cell. Note that the sole computations needed to get this mapping reduces to lines 15 and 17 of \cref{code:spmat}. The rest and majority of the code concerns the creation of arrays to build the corresponding sparse matrices. 

\begin{table}[t]
\begin{center}
\ra{1.2}
\setlength{\tabcolsep}{10pt}
\begin{tabular}{@{}l c@{}}
\toprule
  Architecture \& Training hyparameters  &  mIoU\%
\\
\midrule
$F=256$ \& $3$-dim $\vec{h}_i$ 
    & 76.1 \\
$F=256$ \& $5$-dim $\vec{h}_i$ 
    & 76.6 \\
$F=384$ \& $5$-dim $\vec{h}_i$      
    & 76.9 \\
$F=384$ \& $5$-dim $\vec{h}_i$ \&  Stoch.~depth  
    & 77.6 \\
\bottomrule
\end{tabular}
\end{center}
\caption{Influence of $F$, input vectors $\vec{h}_i$, and stochastic depth on the performance of \ours on nuScenes. We train and evaluate \ours-48-F backbones on the official train and validation set, respectively. We report the average mIoU\% obtained at the last training epoch of two independent runs.}
\label{tab:feat_stoch_depth_nuscenes}
\end{table}

%
\section{Instance Cutmix and Polarmix on SemanticKITTI}

\pgfplotstableread[row sep=\\,col sep=&]{
    interval			& Bas.	& Bas.Aug	& Bas.Cut\\
    {car}			& 94.8	& 95.8     	& 95.4 \\
    {bicycle}		& 44.2 	& 57.9    	& 51.9 \\
    {motorcycle}		& 65.9 	& 76.1     	& 76.0 \\	
    {truck}			& 79.4 	& 82.4     	& 73.9 \\
    {other-vehicle}	& 33.4  	& 52.7    	& 48.1 \\
    {person}		& 63.8 	& 78.2     	& 75.1 \\
    {bicyclist}		& 83.3 	& 93.0     	& 92.9 \\
    {motorcyclist}	& 0.0   	& 0.2	    	&   0.2 \\	
    {road}			& 95.0 	& 95.3     	& 94.7 \\
    {parking}		& 47.5 	& 50.7     	& 49.5 \\	
    {sidewalk}		& 83.2 	& 83.5     	& 82.7 \\	
    {other-ground}	&   0.0  	&  3.2	&   0.3 \\
    {building}		& 90.5 	& 91.4     	& 91.2 \\
    {fence}			& 61.0 	& 64.6    	& 62.7 \\
    {vegetation}		& 88.7 	& 87.6     	& 88.2 \\
    {trunk}			& 68.1 	& 69.4     	& 68.8 \\
    {terrain}		& 76.9 	& 73.5    	& 75.5 \\	
    {pole}			& 63.4 	& 63.9     	& 63.8 \\
    {traffic-sign}		& 48.7 	& 50.2   	& 50.5 \\
    mIoU			& 62.5  	& 66.8     	& 65.3 \\
    }\mydata
    
\begin{figure*}[t]
\centering
\begin{tikzpicture}
    \begin{axis}[
            ybar,
            bar width=.15cm,
            width=\textwidth,
            height=.4\textwidth,
            font=\scriptsize,
            tick label style={font=\scriptsize},
            x tick label style={font=\scriptsize, rotate=40},
            y tick label style={font=\scriptsize, rotate=90},
            legend style={at={(0.5,1.01)}, anchor=south,legend columns=-1,draw=none,/tikz/every even column/.append style={column sep=0.5cm}},
            symbolic x coords={car,bicycle,motorcycle,truck,other-vehicle,person,bicyclist,motorcyclist,road,parking,sidewalk,other-ground,building,fence,vegetation,trunk,terrain,pole,traffic-sign,mIoU},
            xtick=data,
            ytick={0,50,100},
            nodes near coords,
            nodes near coords align={vertical},
            every node near coord/.append style={rotate=90, anchor=west},
            ymin=0,ymax=100,
            enlarge x limits={0.03},
            enlarge y limits={upper,value=0.1},
            xtick pos=left,
            ytick pos=left,
        ]
        \addplot[color=blue3,fill=blue3] table[x=interval,y=Bas.]{\mydata};
        \addplot[color=blue2,fill=blue2] table[x=interval,y=Bas.Cut]{\mydata};
        \addplot[color=blue2,fill=blue2] table[x=interval,y=Bas.Aug]{\mydata};        
        \legend{Baseline,Baseline + Instance Cutmix,Baseline + Polarmix + Instance Cutmix}
    \end{axis}
\end{tikzpicture}
\caption{Performance of \ours when using baseline augmentations (rotation, flip axis, scaling), when adding instance cutmix, or when adding instance cutmix and polarmix together. We train a \ours-48-256 backbone on the train set of SemanticKITTI and compute the mIoU on the corresponding validation set. We report the average mIoU\% obtained at the last training epoch of two runs.}
\label{fig:polar_cutmix}
\end{figure*}
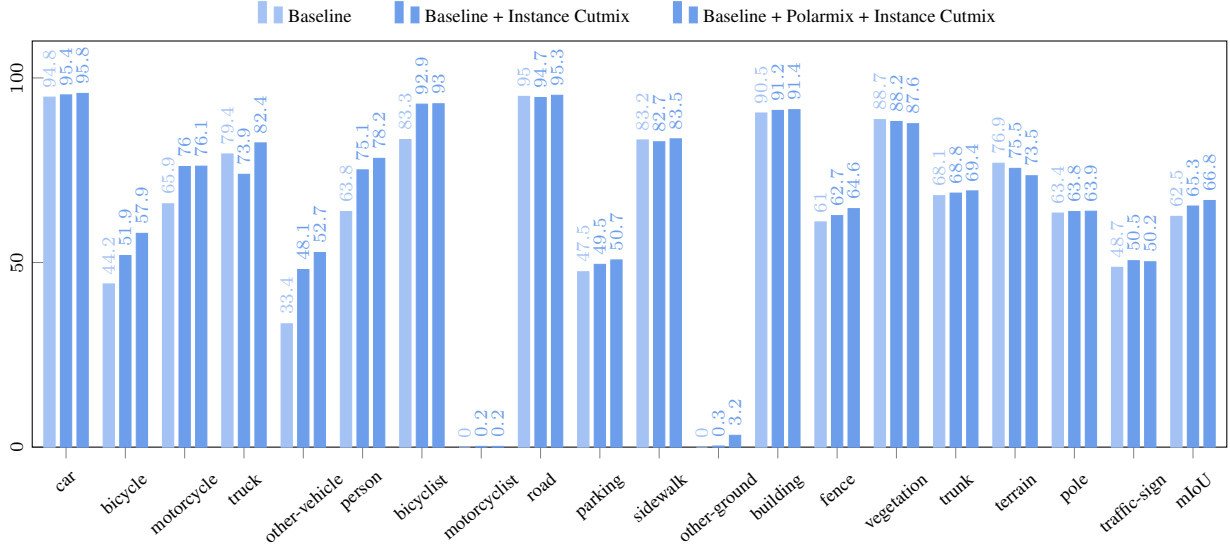

In complement to Sec.~4.5, we show in \cref{fig:polar_cutmix} the benefit of combining the augmentations polarmix \cite{polarmix} and instance cutmix \cite{2dpass,RPVNet} over instance cutmix alone for training on SemanticKITTI \cite{semkitti}. The combination allows to us to improve the mIoU\% by 1.5 point on average, with the most notable improvements in the classes bicycle, other-vehicle and person.

%
\section{Input Features, dimension $F$, and Stochastic Depth on nuScenes}

We present in \cref{tab:feat_stoch_depth_nuscenes} the interest of successively: using $5$-dimensional input vectors $\vec{h}_i$ (intensity, $x$, $y$, $z$, and range of $\vec{p}_i$) instead of $3$-dimensional input vectors (intensity, height=$z$, range); increasing $F$ from 256 to 384; and using stochastic depth during training on nuScenes. We notice that each of these ingredients improves the mIoU\% to finally reach $77.6$ on average over two independent training.

%
\section{Visual Inspections}

\begin{figure*}[h]
\begin{center}
\begin{minipage}{0.31\linewidth}
\centering \small Ground truth
\end{minipage}
\begin{minipage}{0.31\linewidth}
\centering \small \ours's result
\end{minipage}
\begin{minipage}{0.31\linewidth}
\centering \small Wrong classifications in red
\end{minipage}
\vspace{3mm}
\\
\includegraphics[width=0.31\linewidth,trim={5cm 1cm 5cm 2cm},clip]{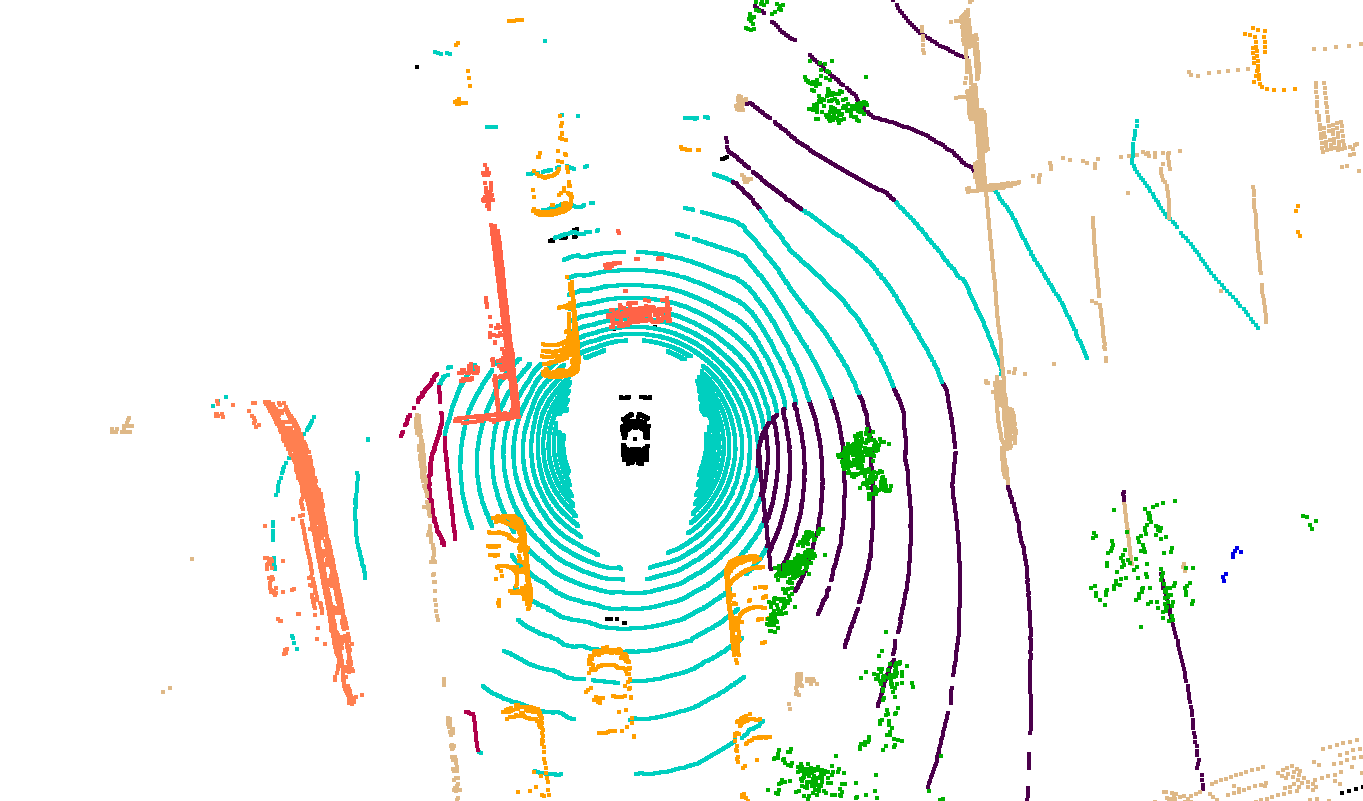}
\includegraphics[width=0.31\linewidth,trim={5cm 1cm 5cm 2cm},clip]{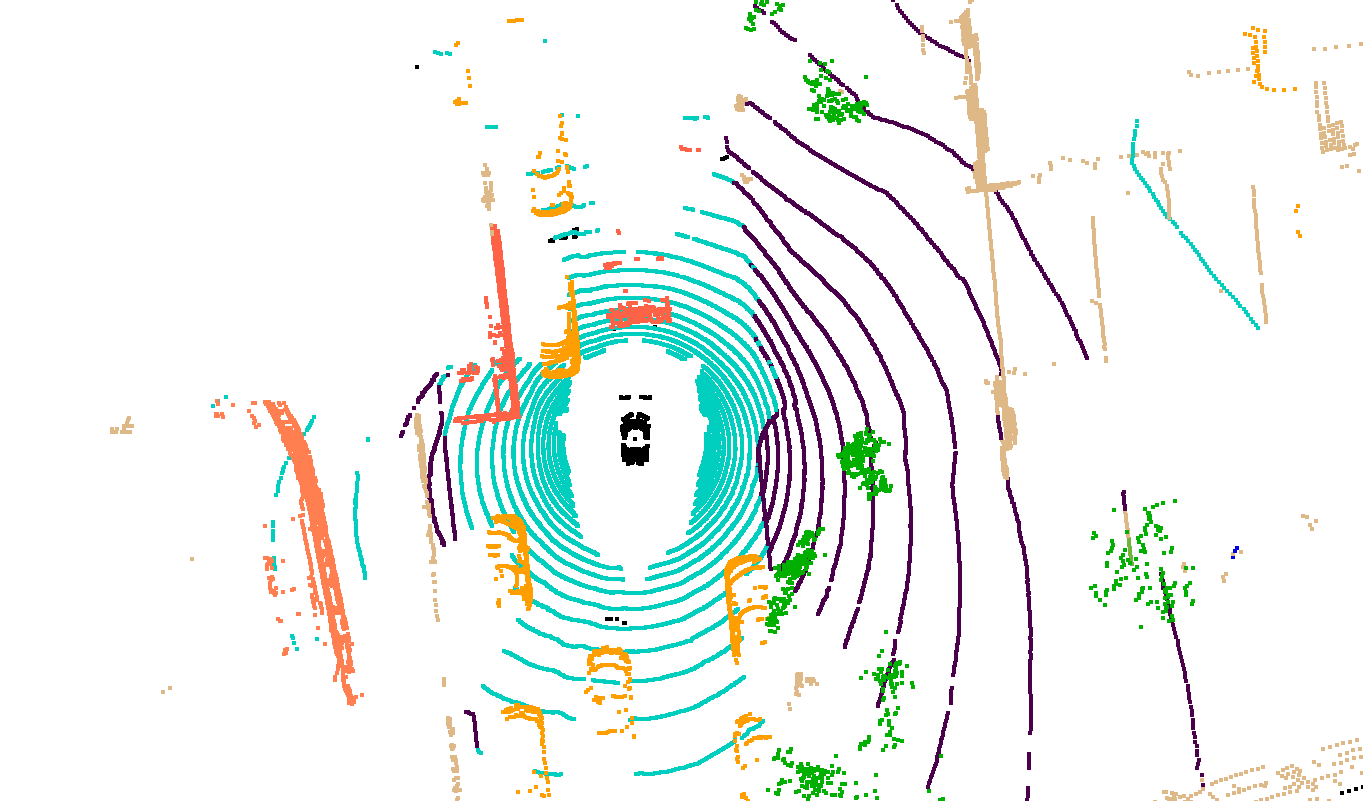}
\includegraphics[width=0.31\linewidth,trim={5cm 1cm 5cm 2cm},clip]{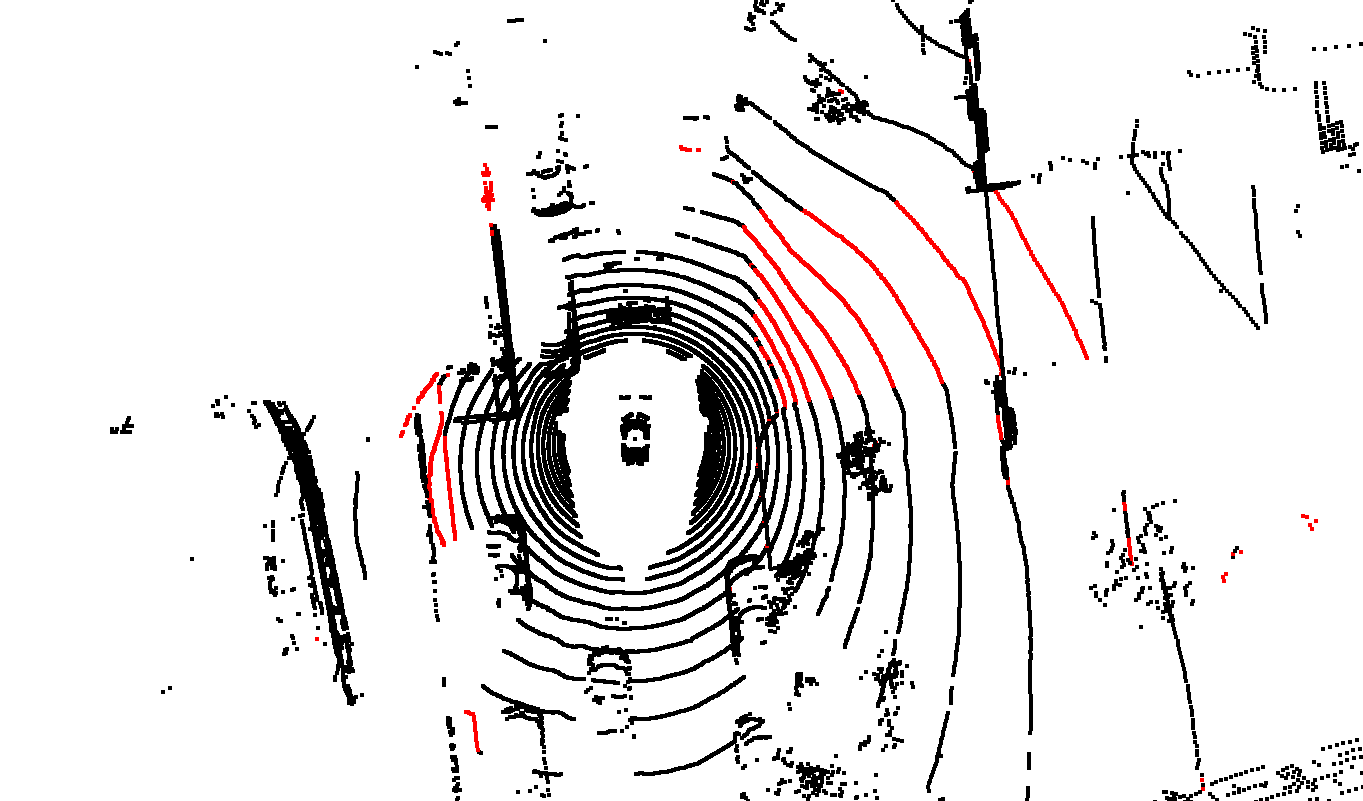}
\\
\includegraphics[width=0.31\linewidth,trim={8cm 3cm 10cm 5cm},clip]{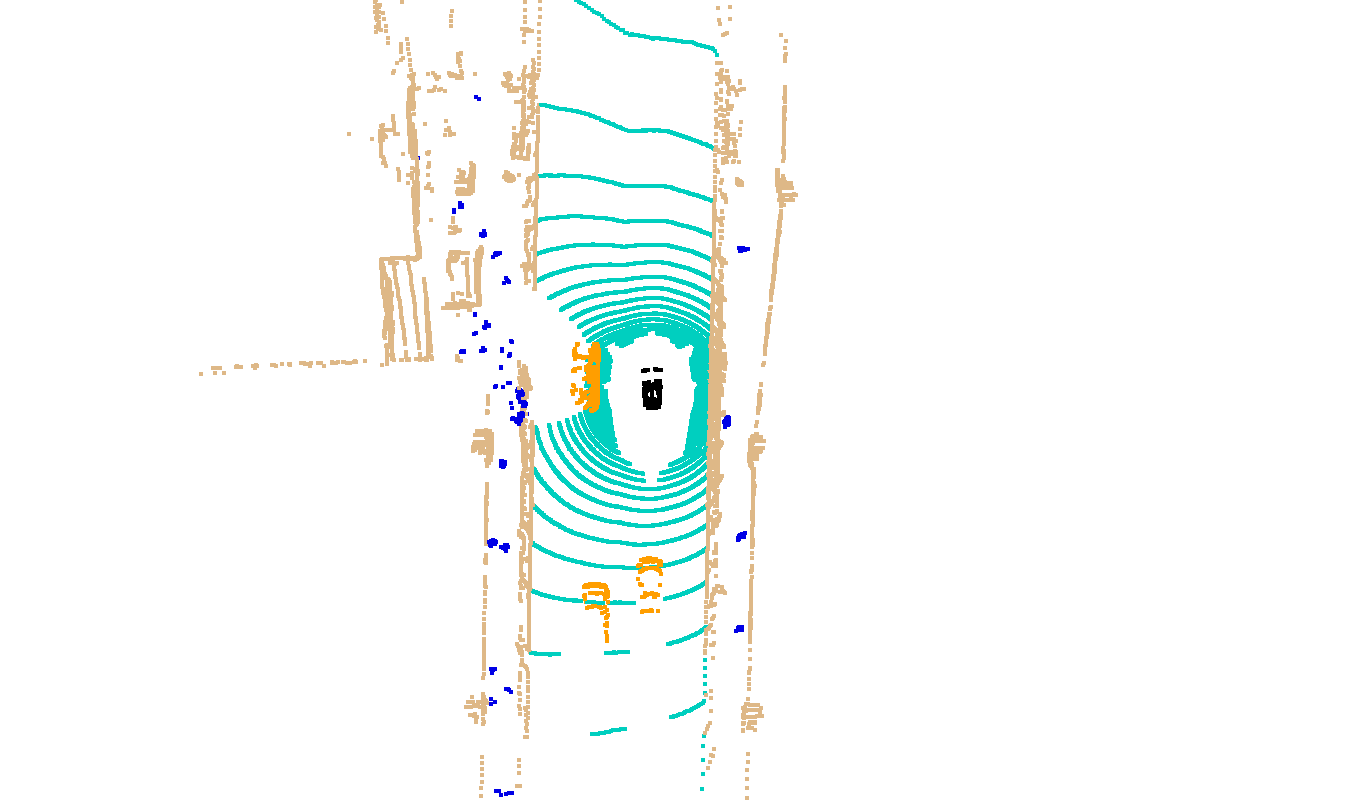}
\includegraphics[width=0.31\linewidth,trim={8cm 3cm 10cm 5cm},clip]{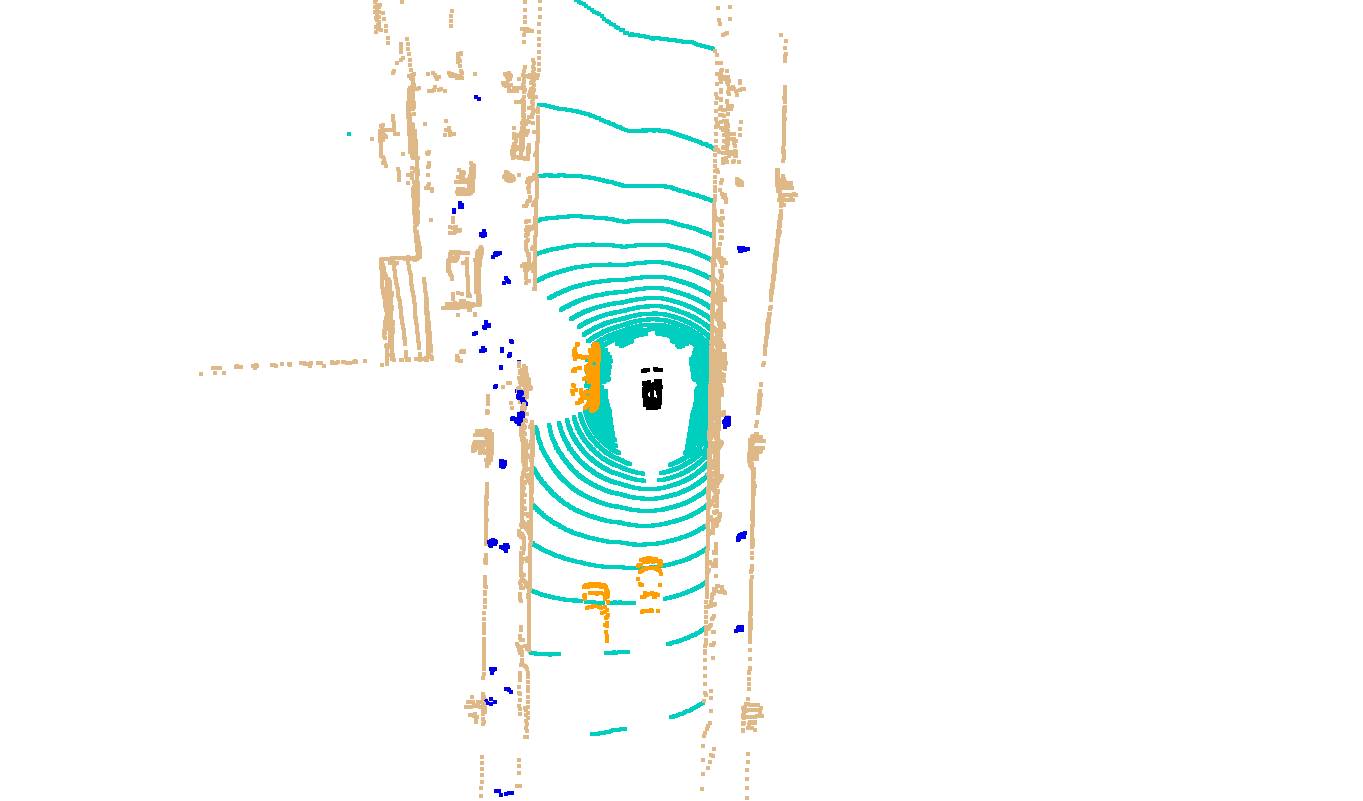}
\includegraphics[width=0.31\linewidth,trim={5cm 3cm 10cm 5cm},clip]{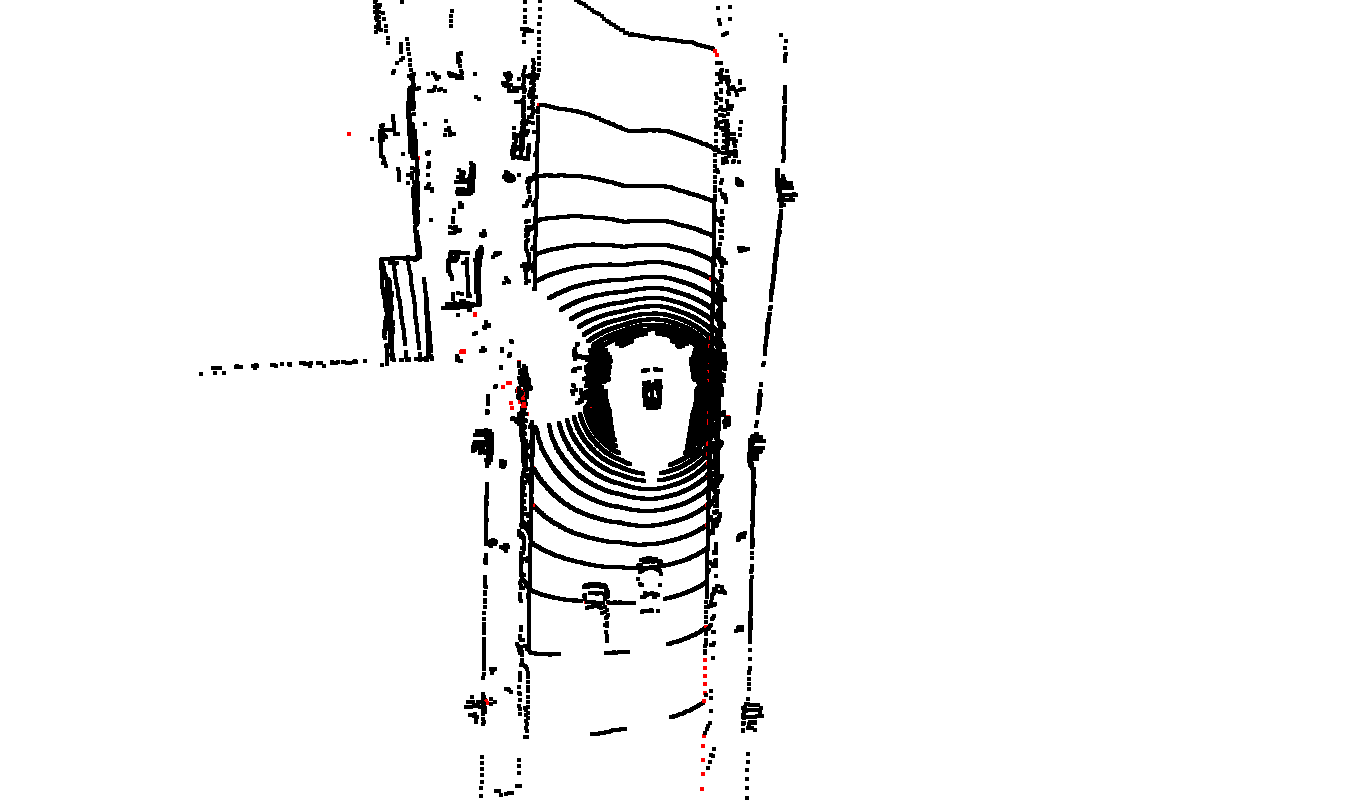}
\\
\includegraphics[width=0.31\linewidth,trim={10cm 1cm 10cm 5cm},clip]{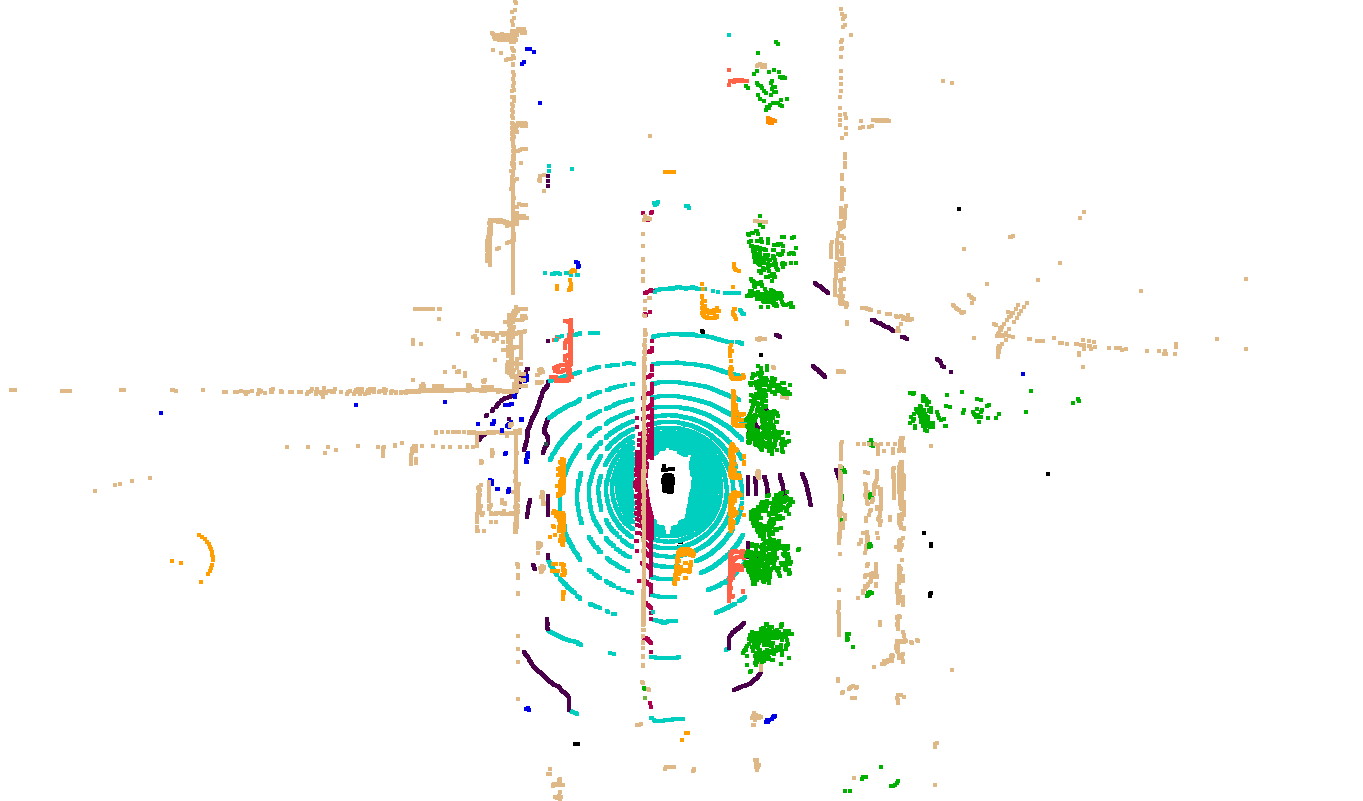}
\includegraphics[width=0.31\linewidth,trim={10cm 1cm 10cm 5cm},clip]{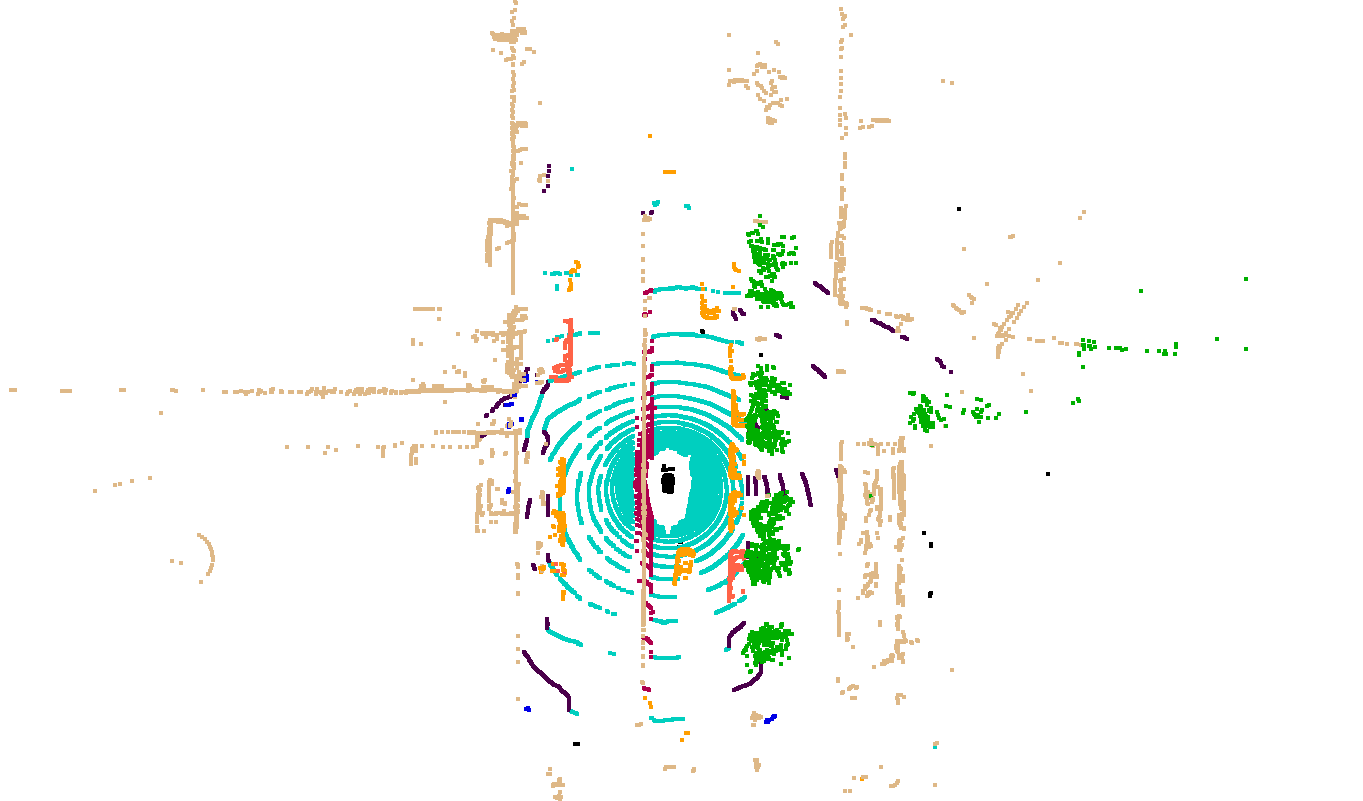}
\includegraphics[width=0.31\linewidth,trim={10cm 1cm 10cm 5cm},clip]{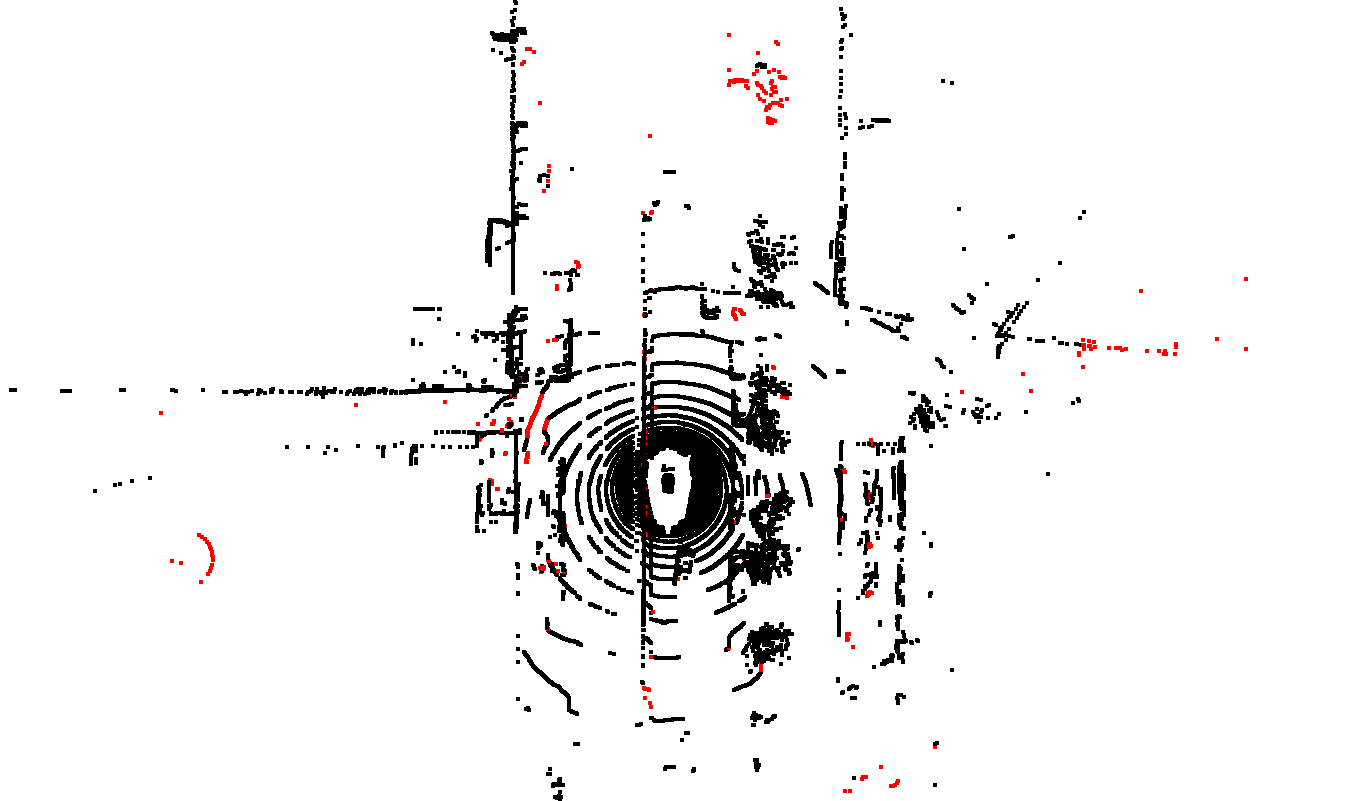}
\\
\includegraphics[width=0.31\linewidth,trim={10cm 1cm 10cm 5cm},clip]{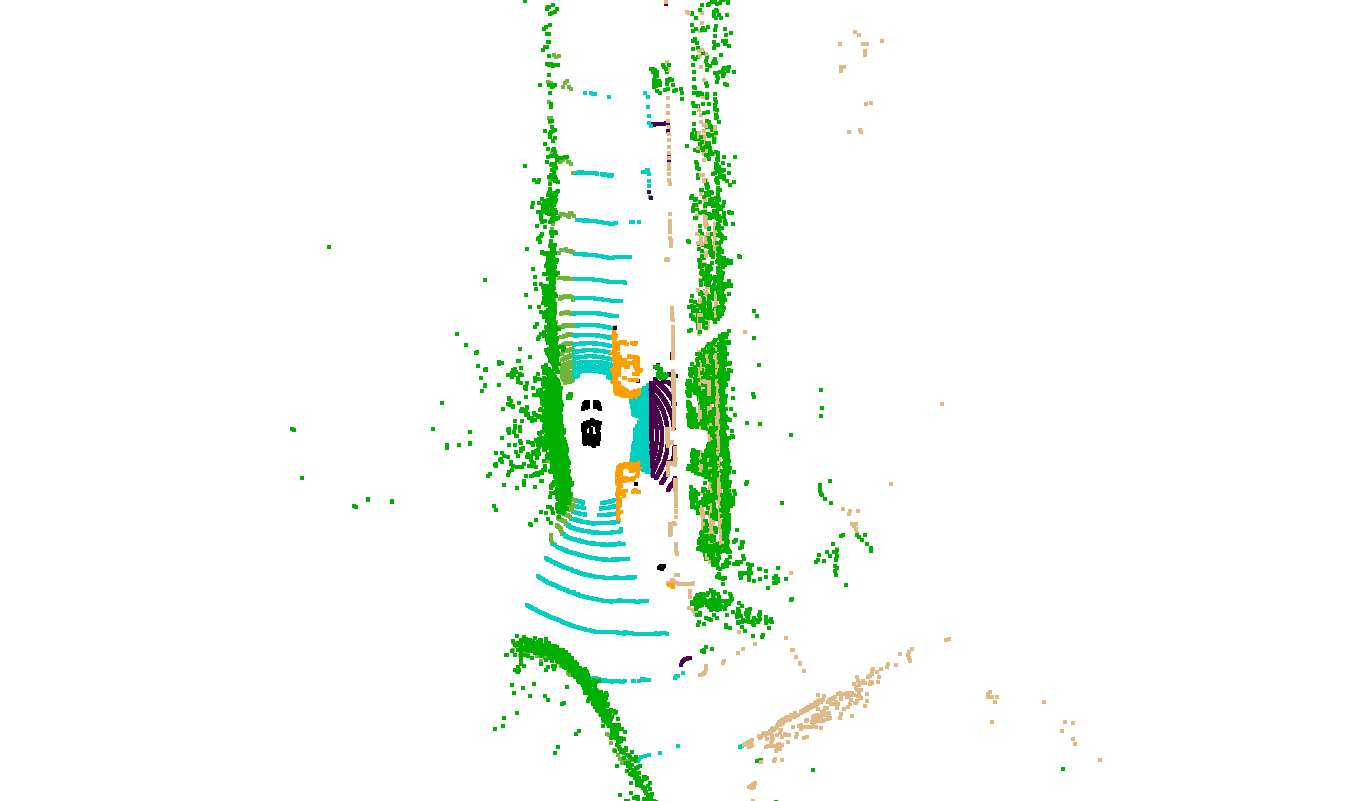}
\includegraphics[width=0.31\linewidth,trim={10cm 1cm 10cm 5cm},clip]{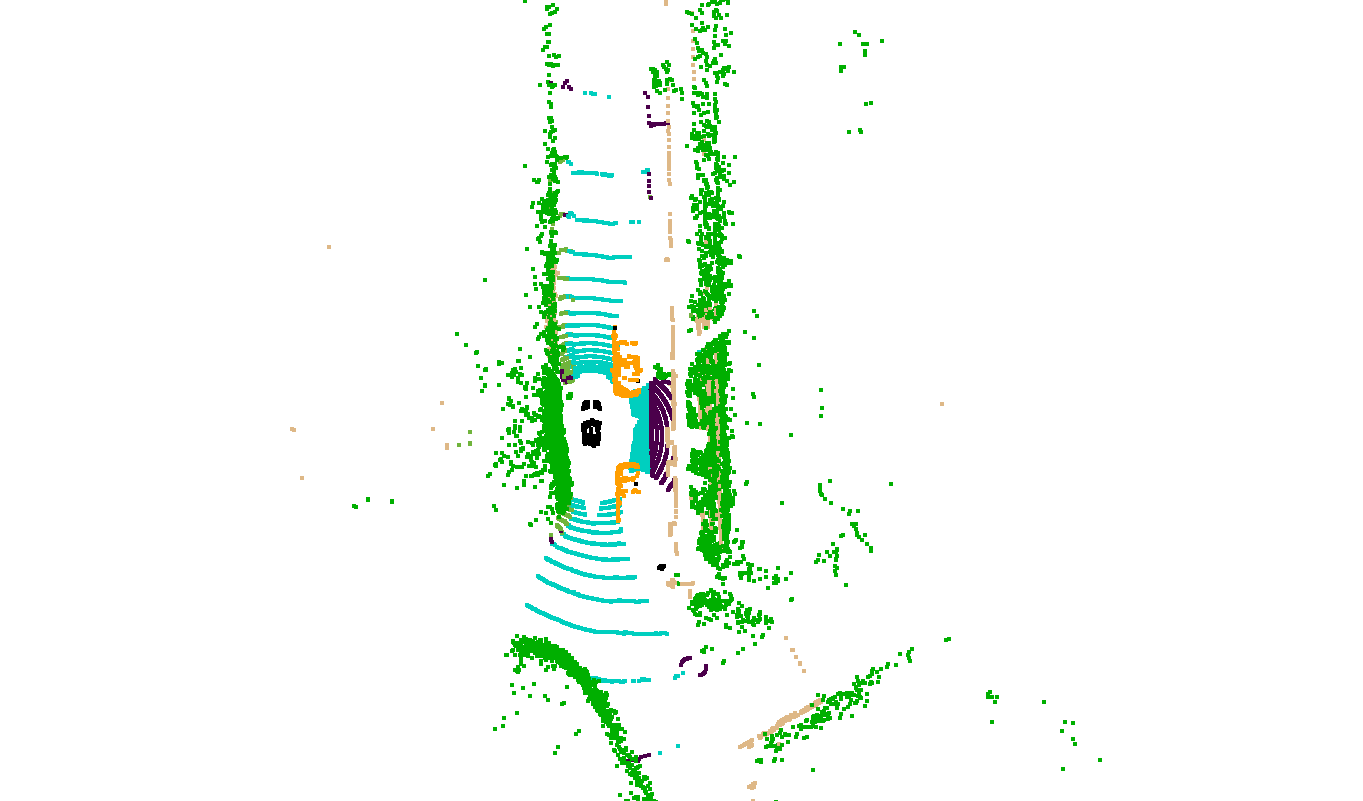}
\includegraphics[width=0.31\linewidth,trim={10cm 1cm 10cm 5cm},clip]{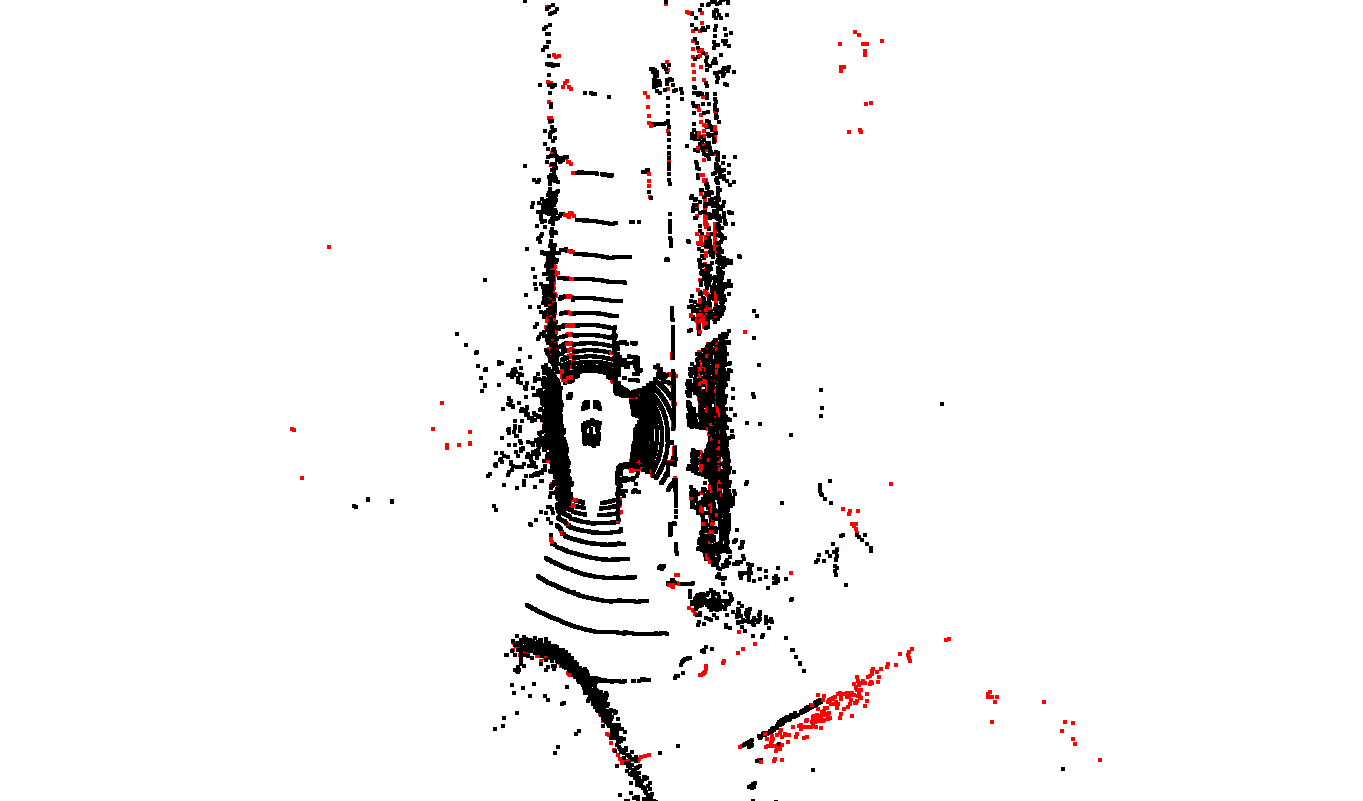}
\\
\caption{Visualization of semantic segmentation results on the validation set of nuScenes obtained with \ours.} 
\label{fig:nuscenes}
\end{center}
\end{figure*}

\begin{figure*}[h]
\begin{center}
\begin{minipage}{0.31\linewidth}
\centering \small Ground truth
\end{minipage}
\begin{minipage}{0.31\linewidth}
\centering \small \ours's result
\end{minipage}
\begin{minipage}{0.31\linewidth}
\centering \small Wrong classifications in red
\end{minipage}
\vspace{3mm}
\\
\includegraphics[width=0.31\linewidth,trim={5cm 5cm 5cm 0},clip]{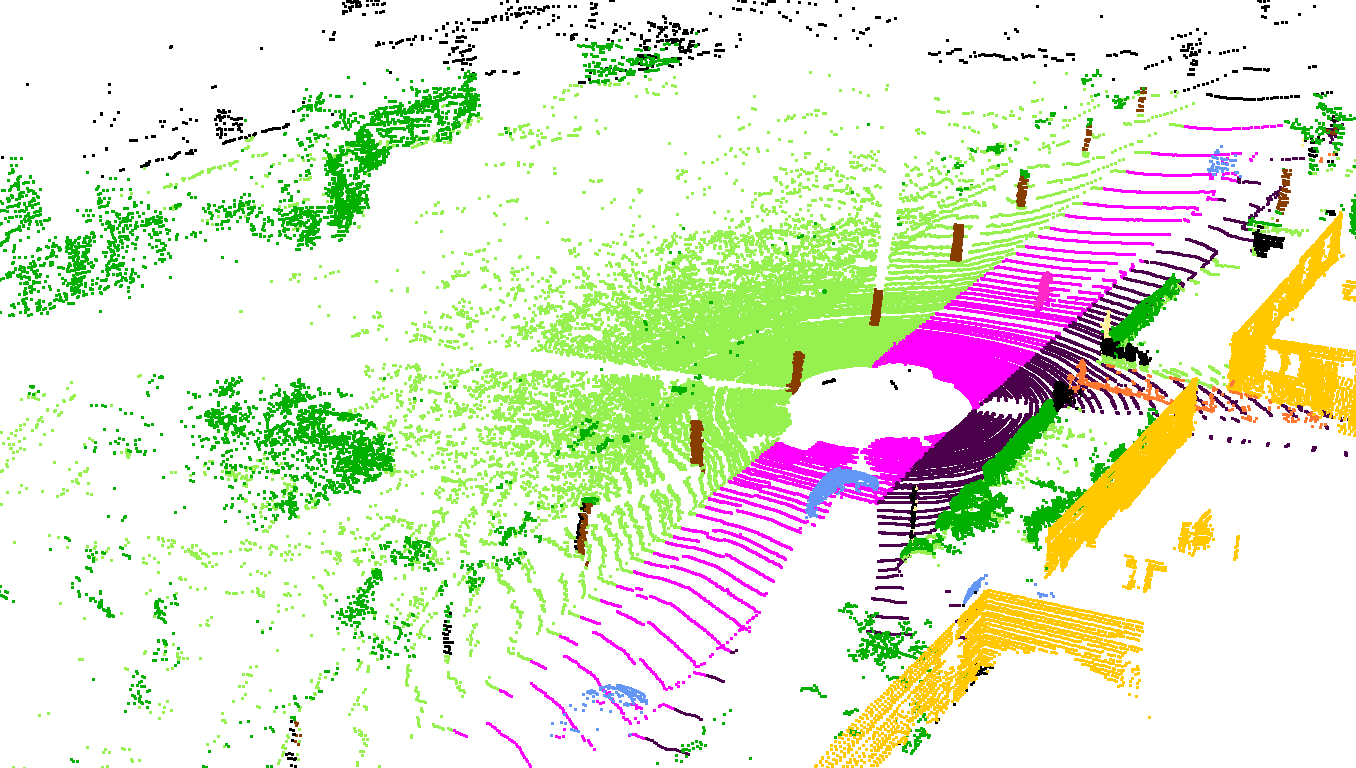}
\includegraphics[width=0.31\linewidth,trim={5cm 5cm 5cm 0},clip]{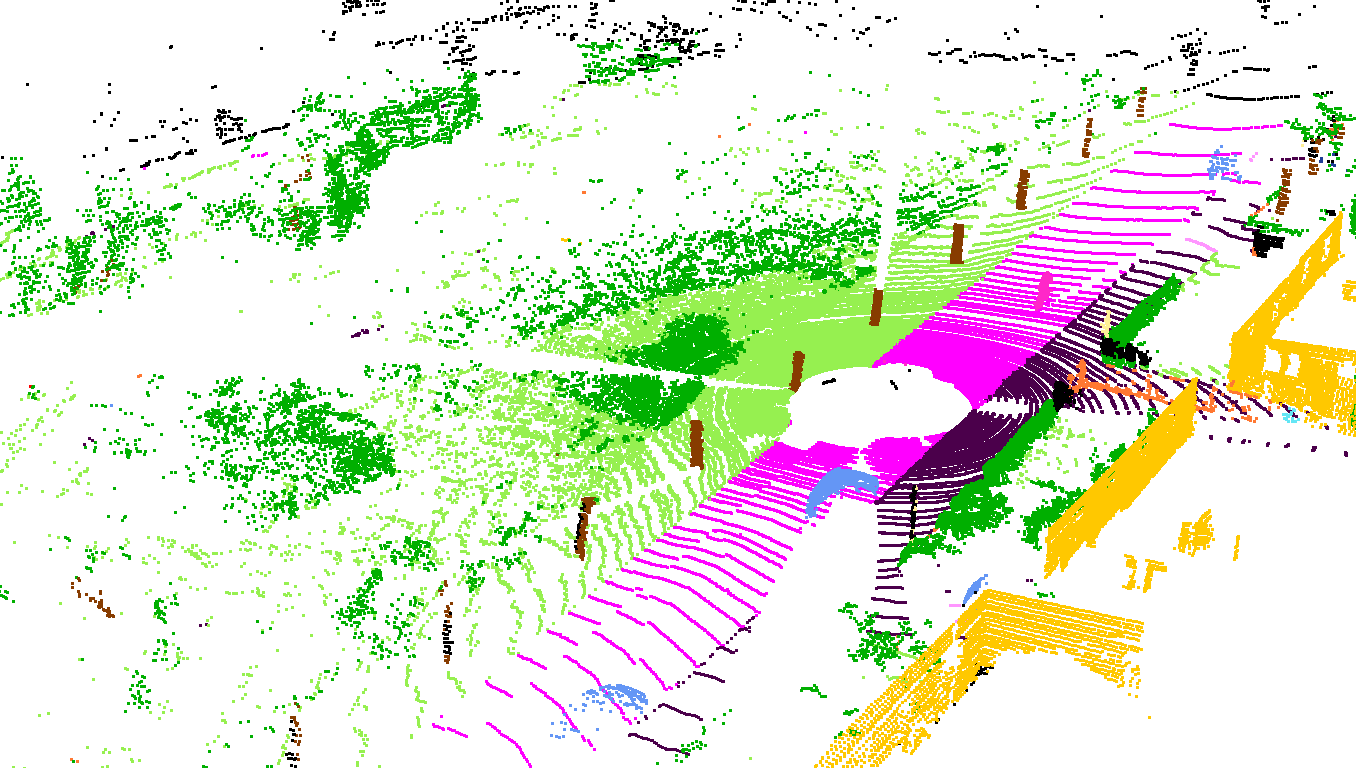}
\includegraphics[width=0.31\linewidth,trim={5cm 5cm 5cm 0},clip]{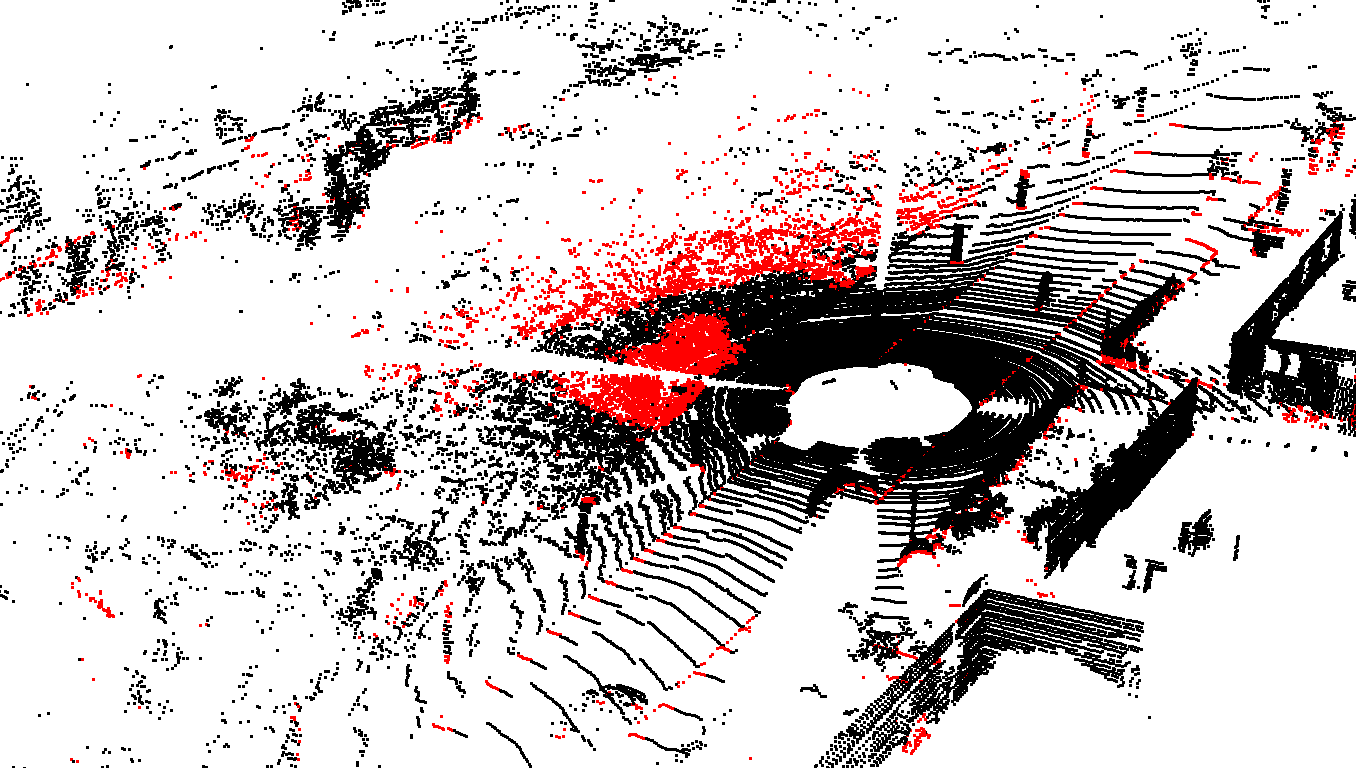}
\\
\includegraphics[width=0.31\linewidth,trim={5cm 0 5cm 0},clip]{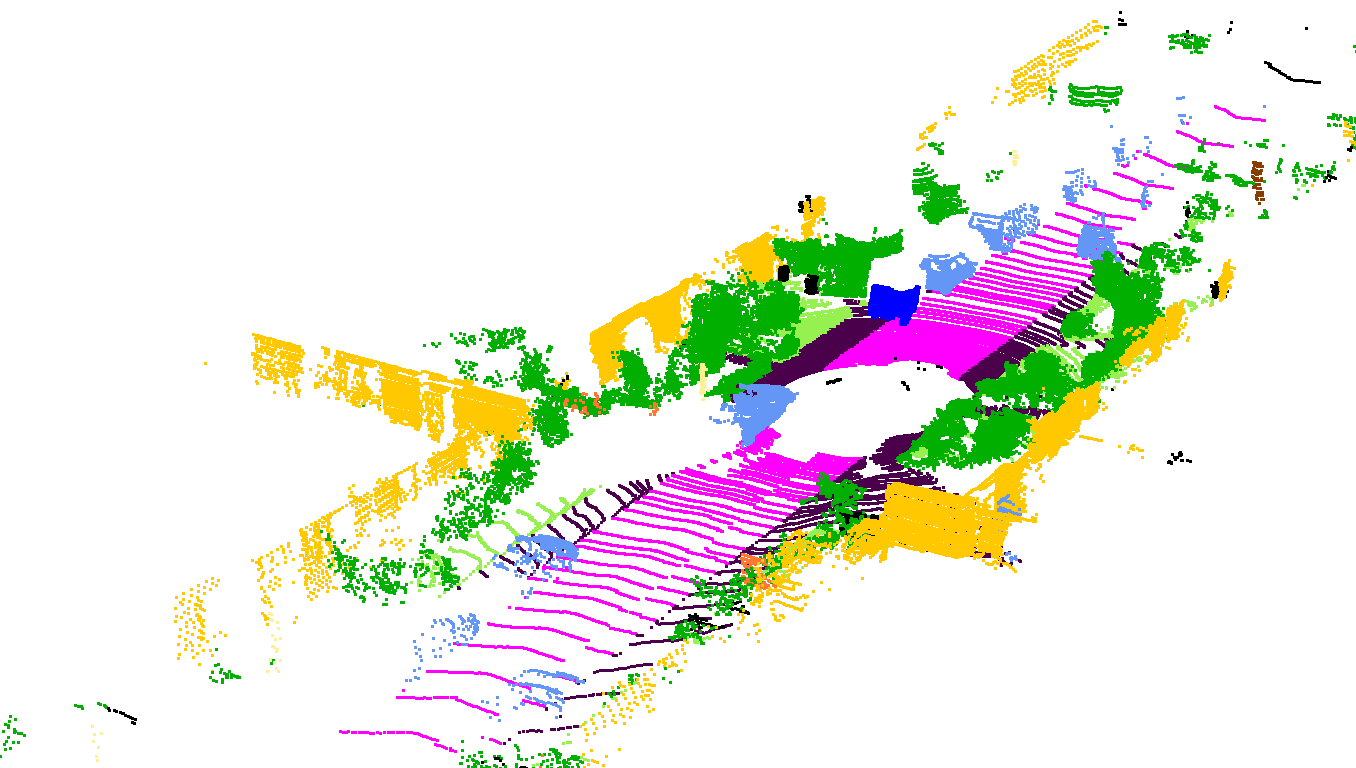}
\includegraphics[width=0.31\linewidth,trim={5cm 0 5cm 0},clip]{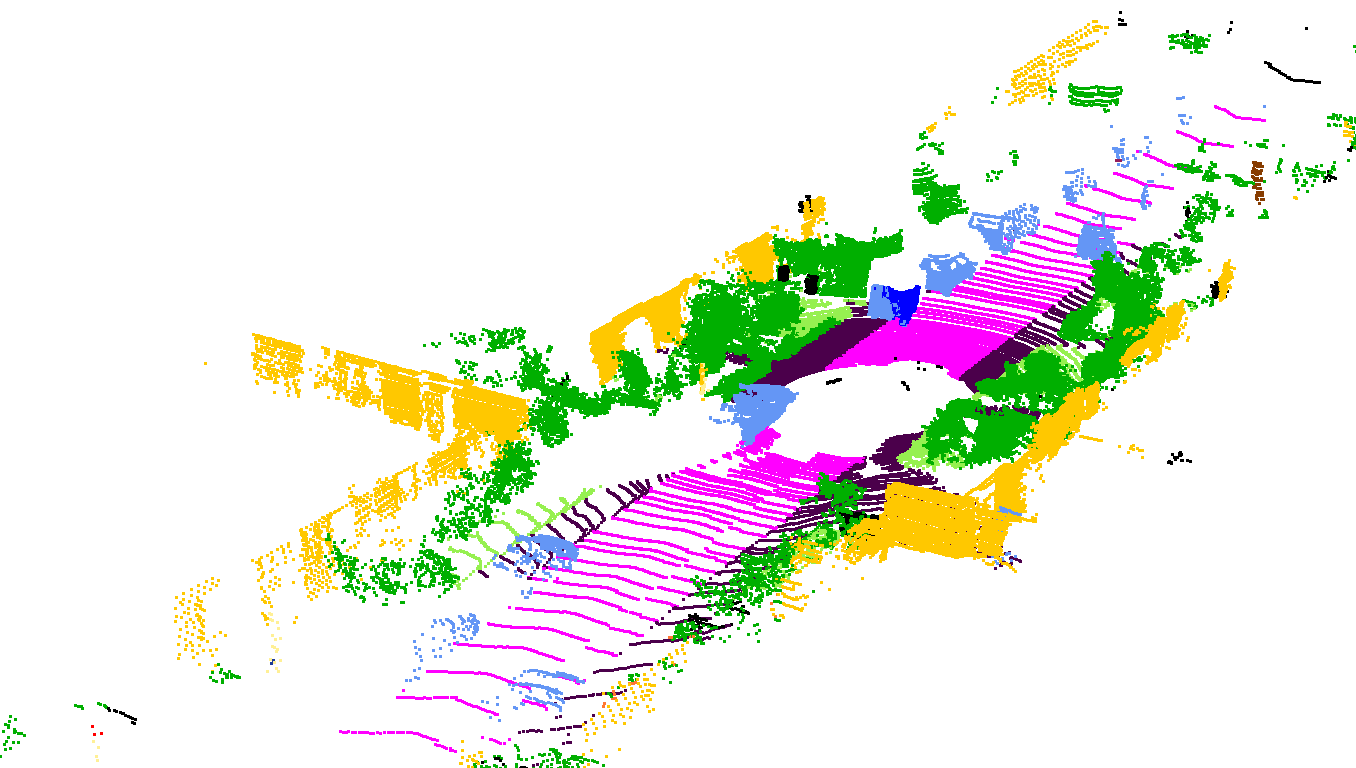}
\includegraphics[width=0.31\linewidth,trim={5cm 0 5cm 0},clip]{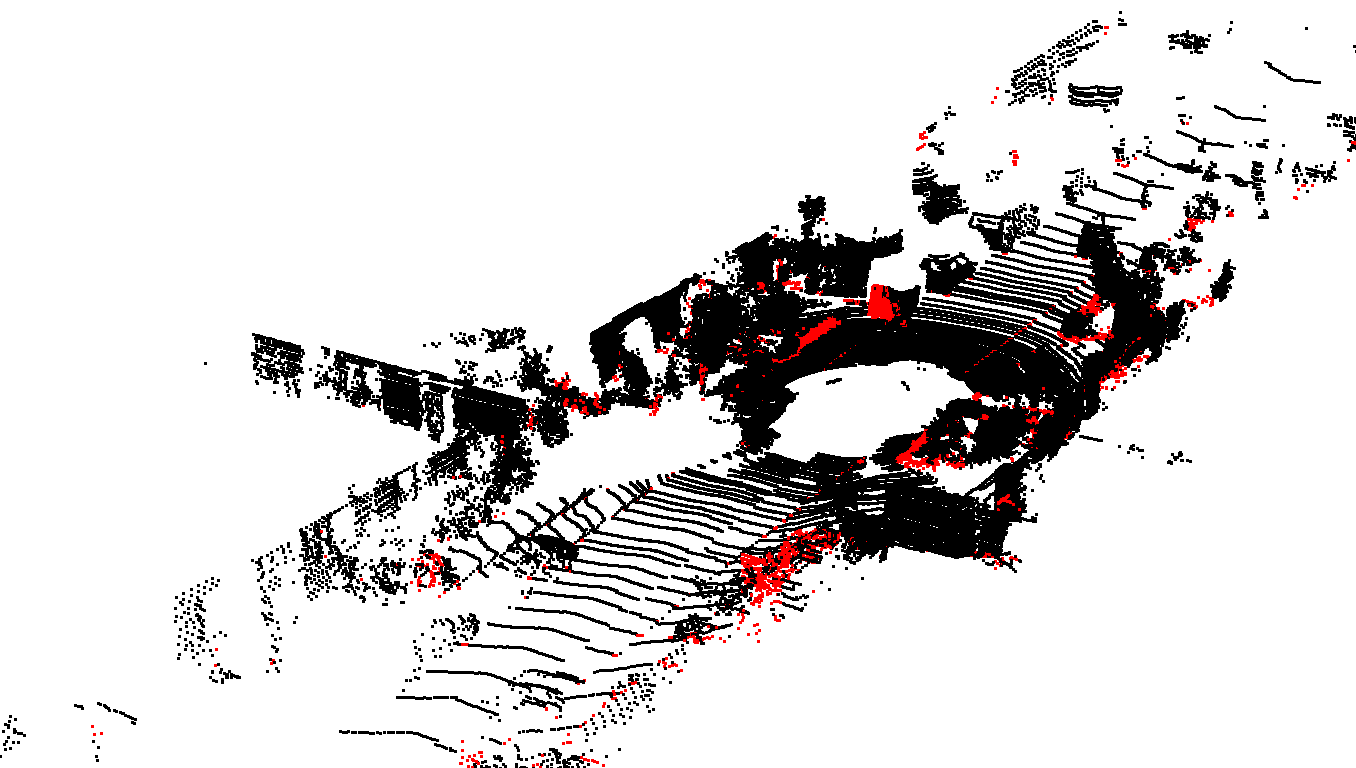}
\\
\includegraphics[width=0.31\linewidth,trim={5cm 0 5cm 2cm},clip]{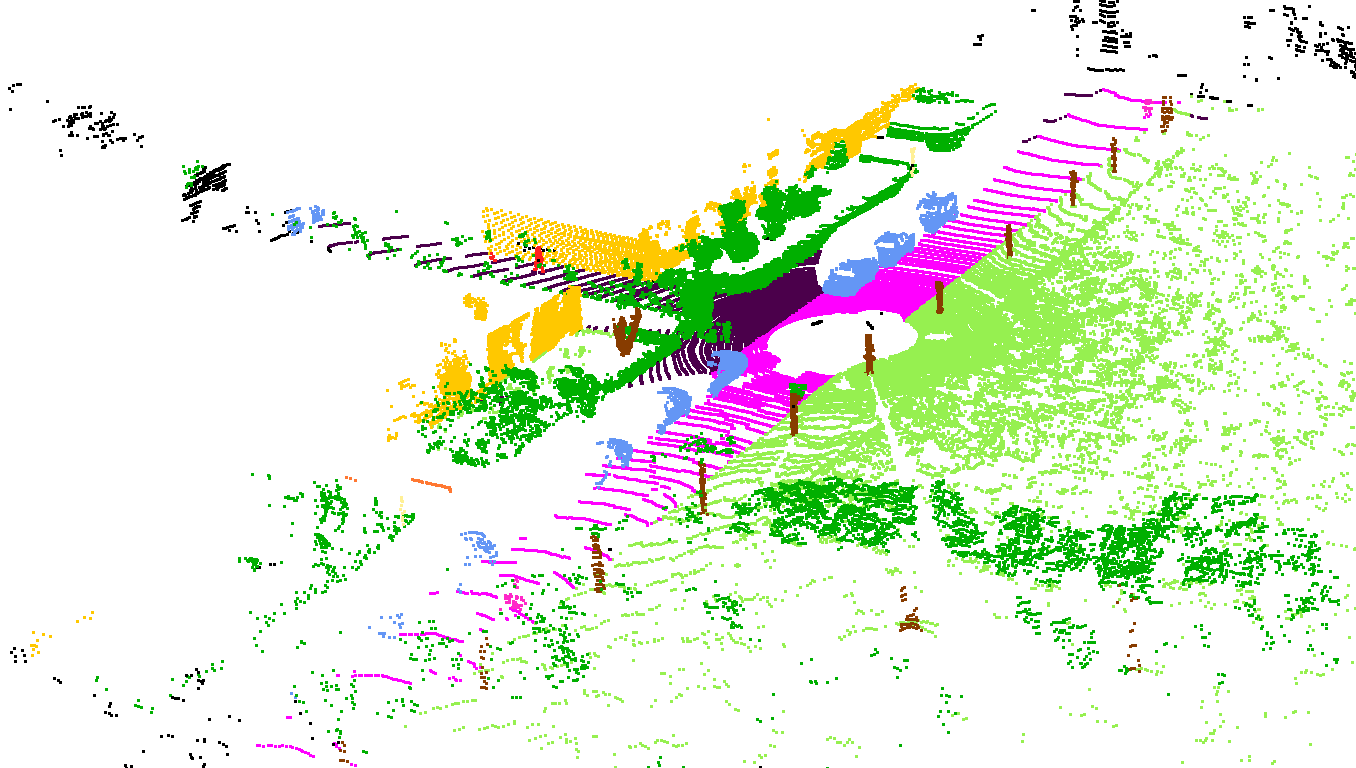}
\includegraphics[width=0.31\linewidth,trim={5cm 0 5cm 2cm},clip]{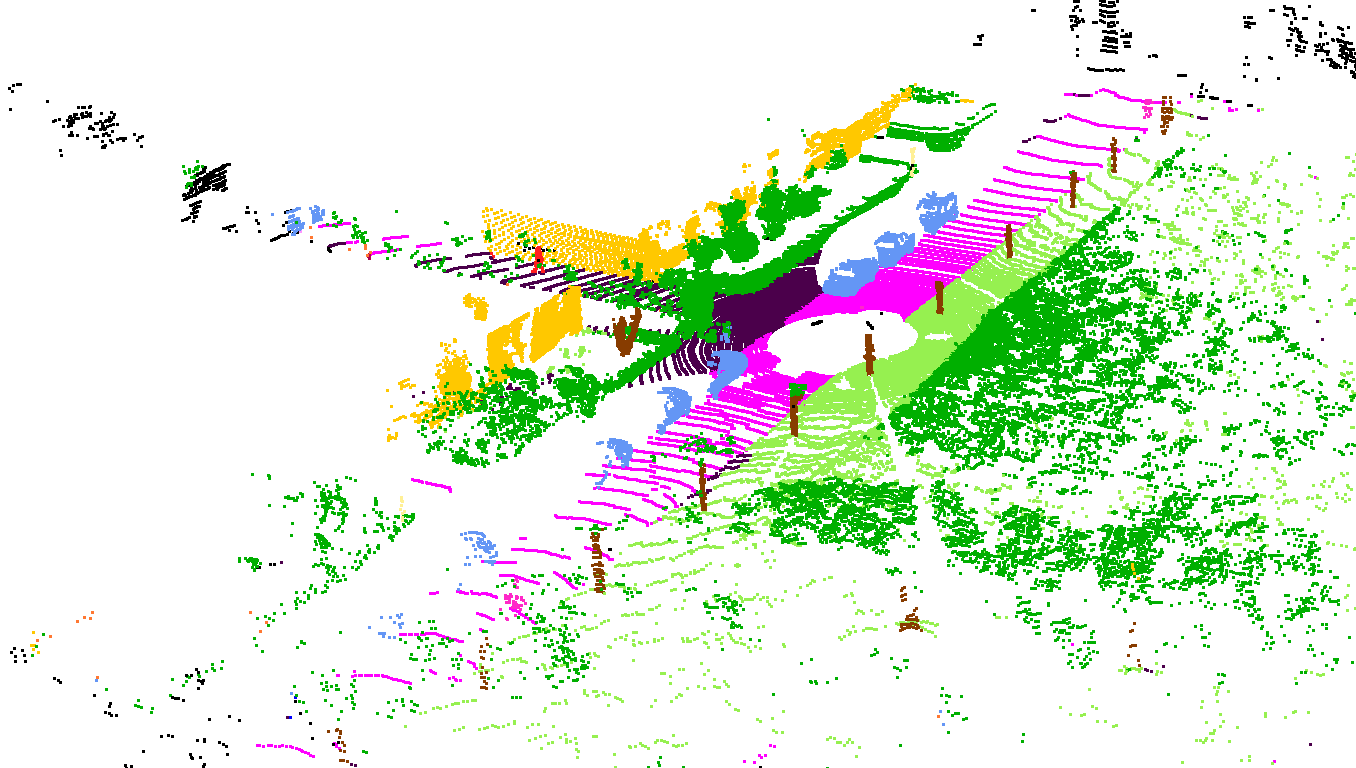}
\includegraphics[width=0.31\linewidth,trim={5cm 0 5cm 2cm},clip]{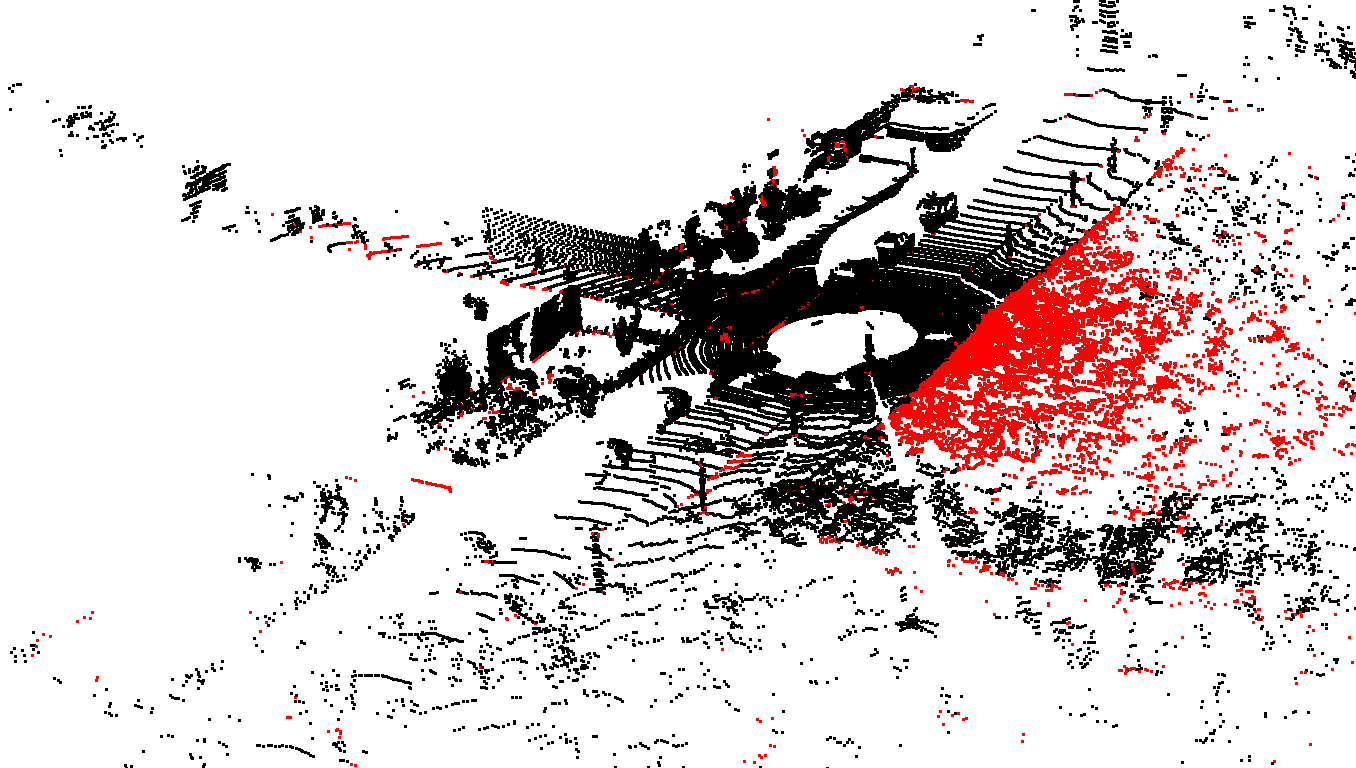}
\\
\includegraphics[width=0.31\linewidth,trim={5cm 0cm 8cm 5cm},clip]{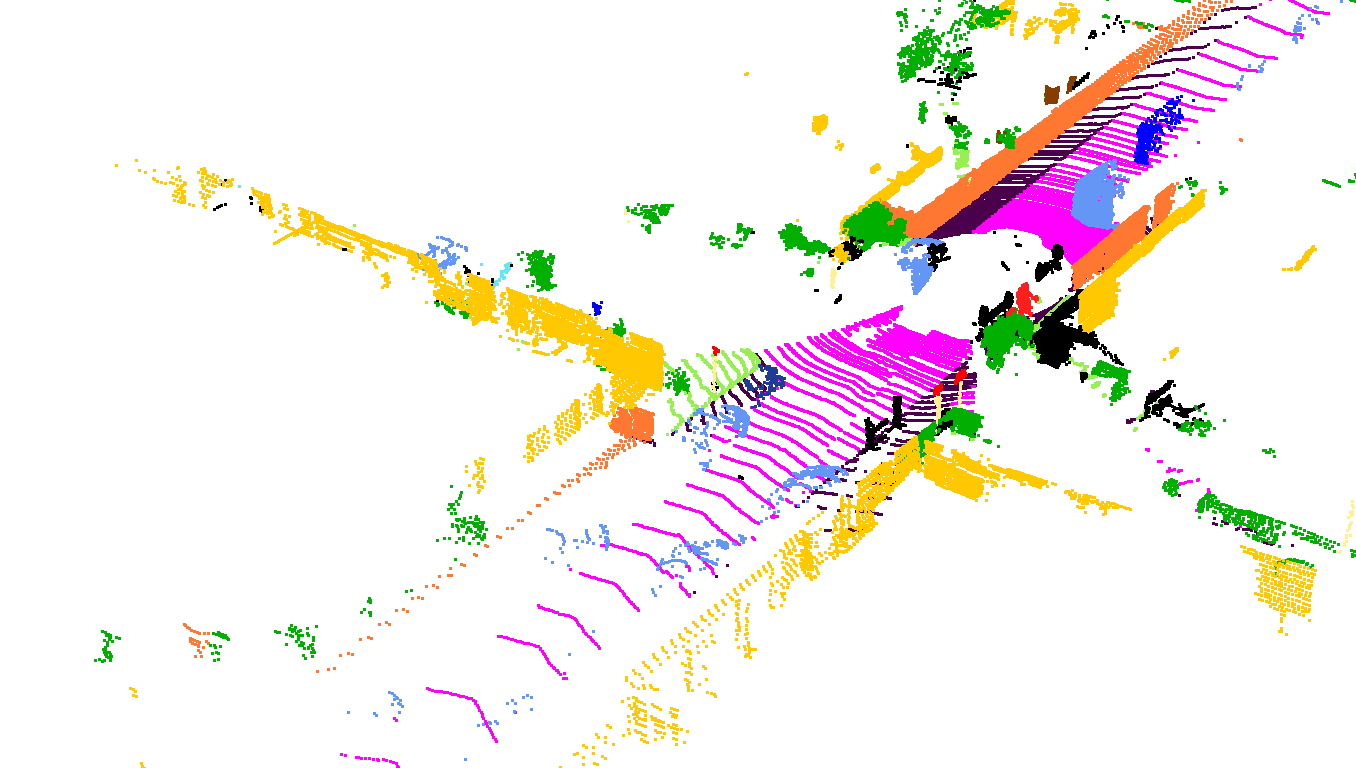}
\includegraphics[width=0.31\linewidth,trim={5cm 0cm 8cm 5cm},clip]{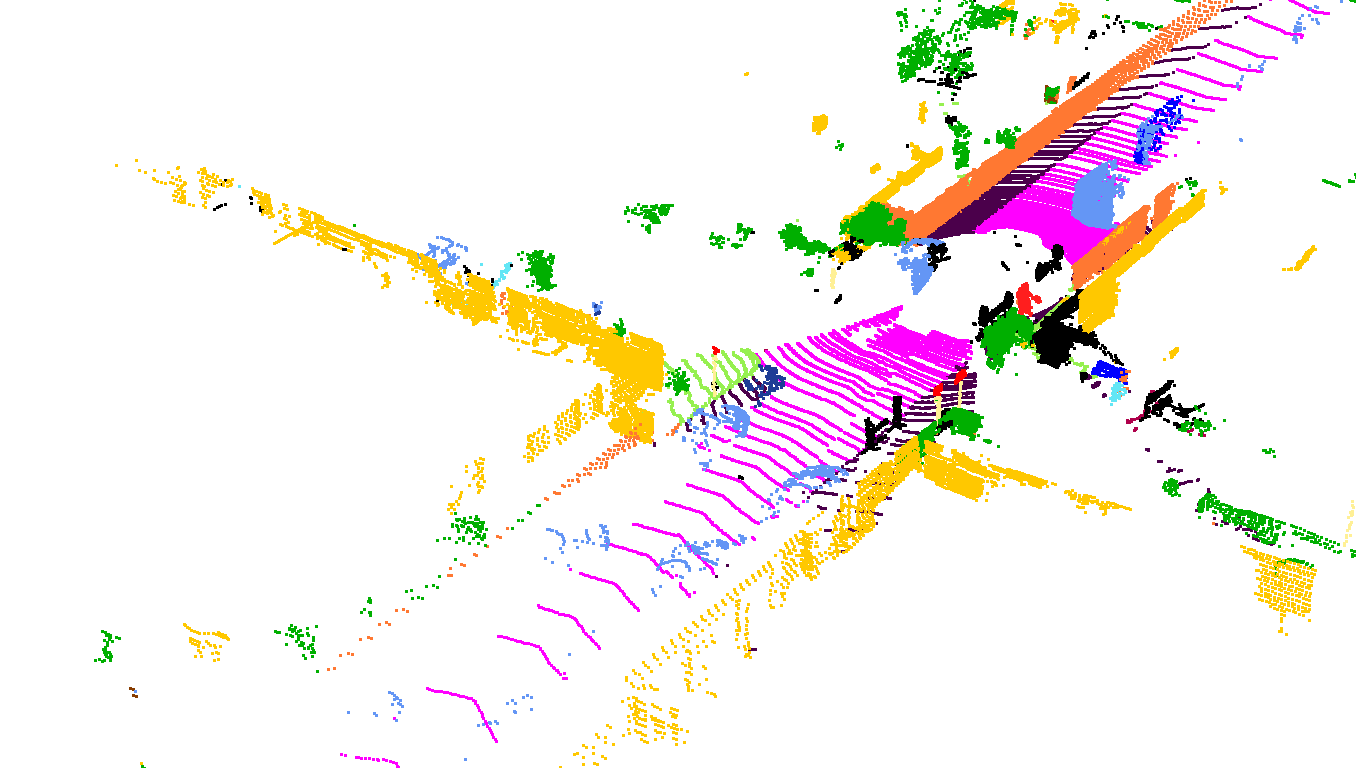}
\includegraphics[width=0.31\linewidth,trim={5cm 0cm 8cm 5cm},clip]{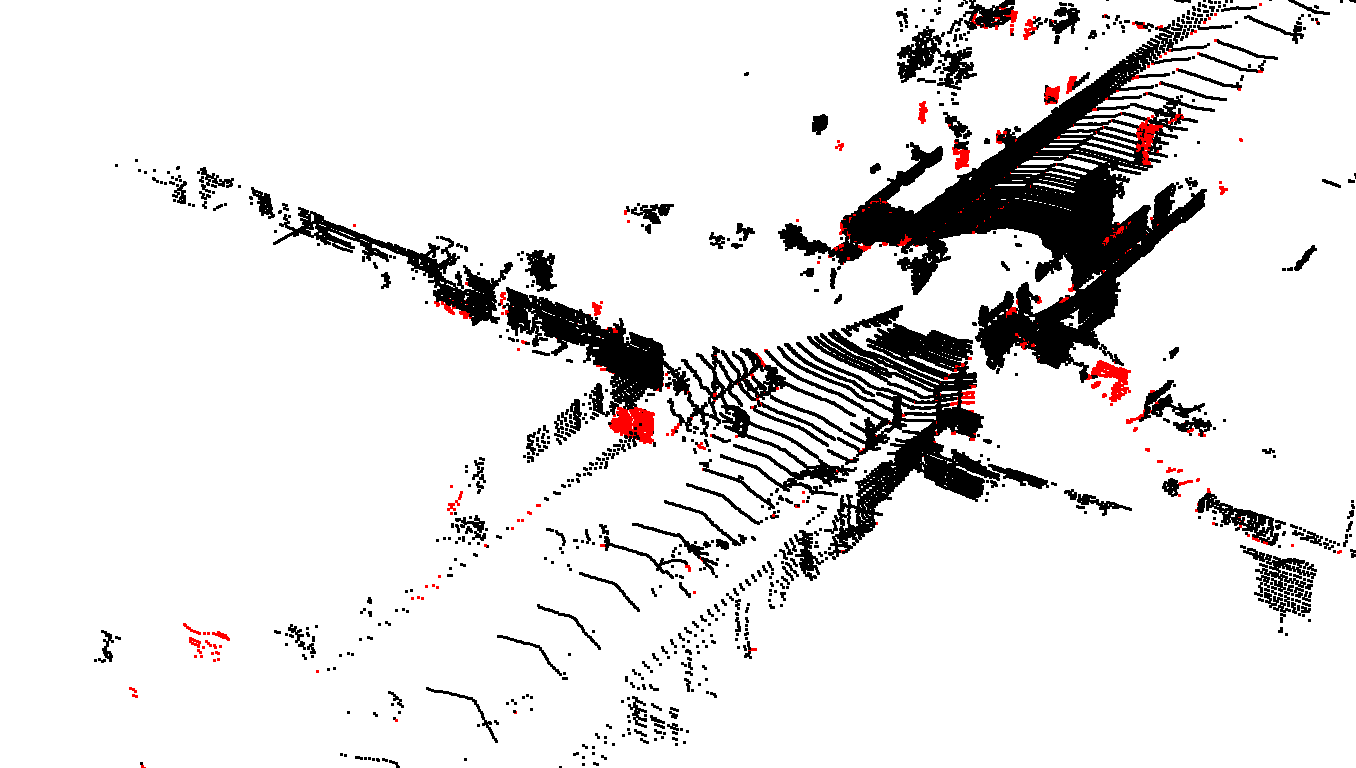}
\\
\includegraphics[width=0.31\linewidth,trim={10cm 0cm 0 2cm},clip]{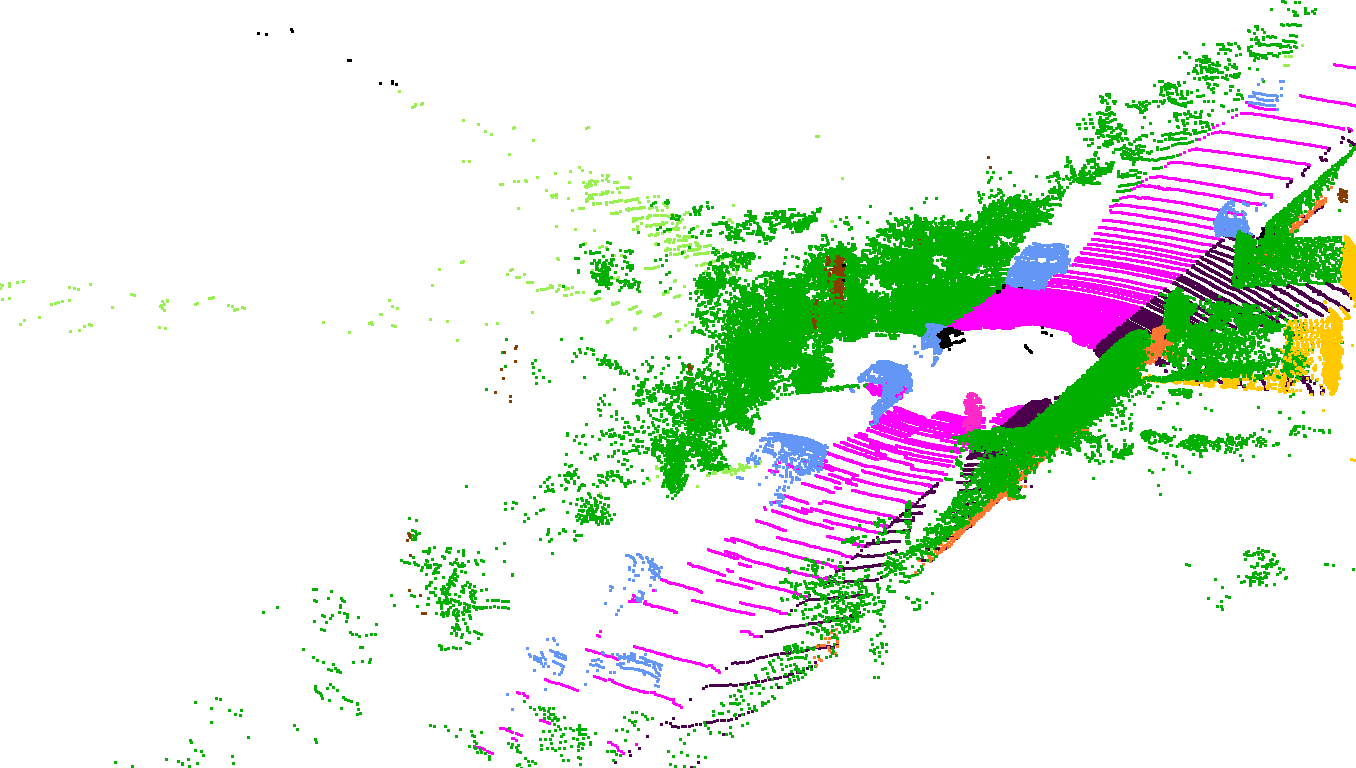}
\includegraphics[width=0.31\linewidth,trim={10cm 0 0 2cm},clip]{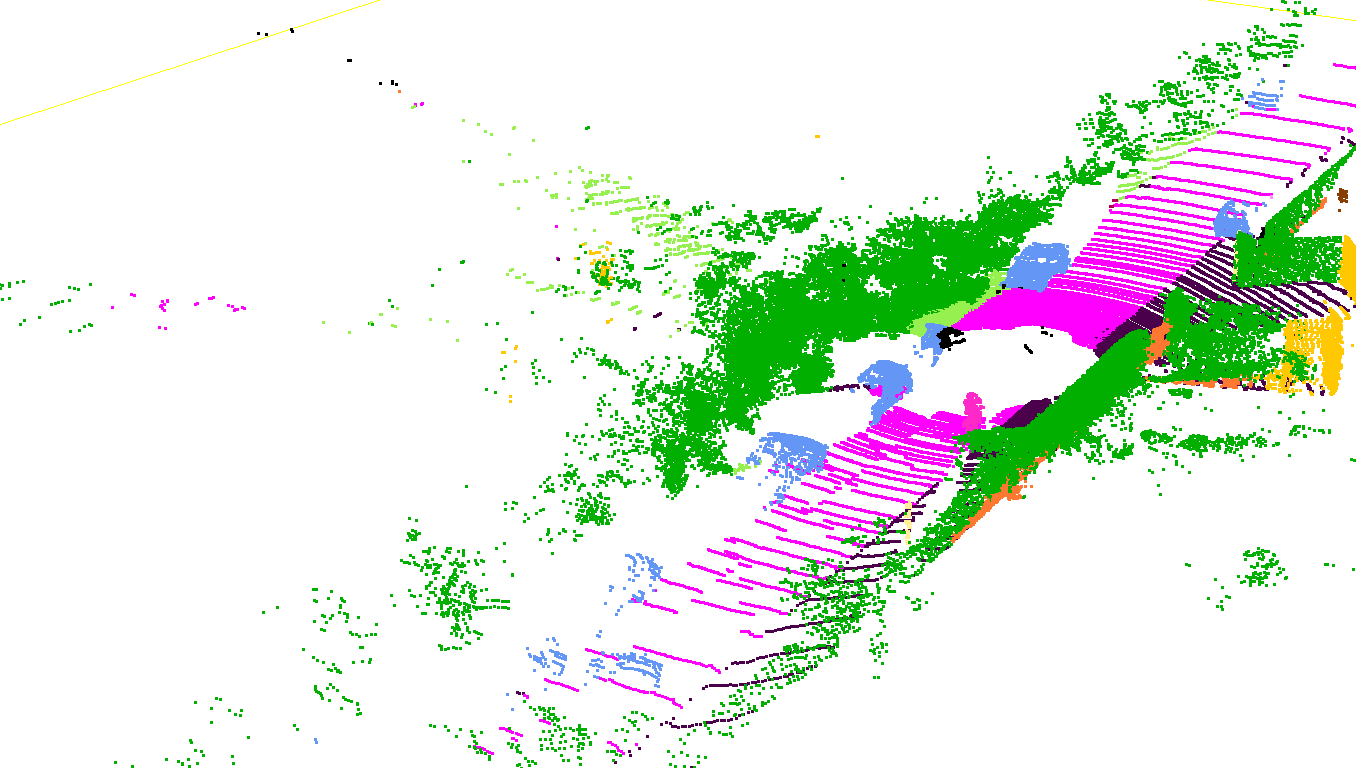}
\includegraphics[width=0.31\linewidth,trim={10cm 0 0 2cm},clip]{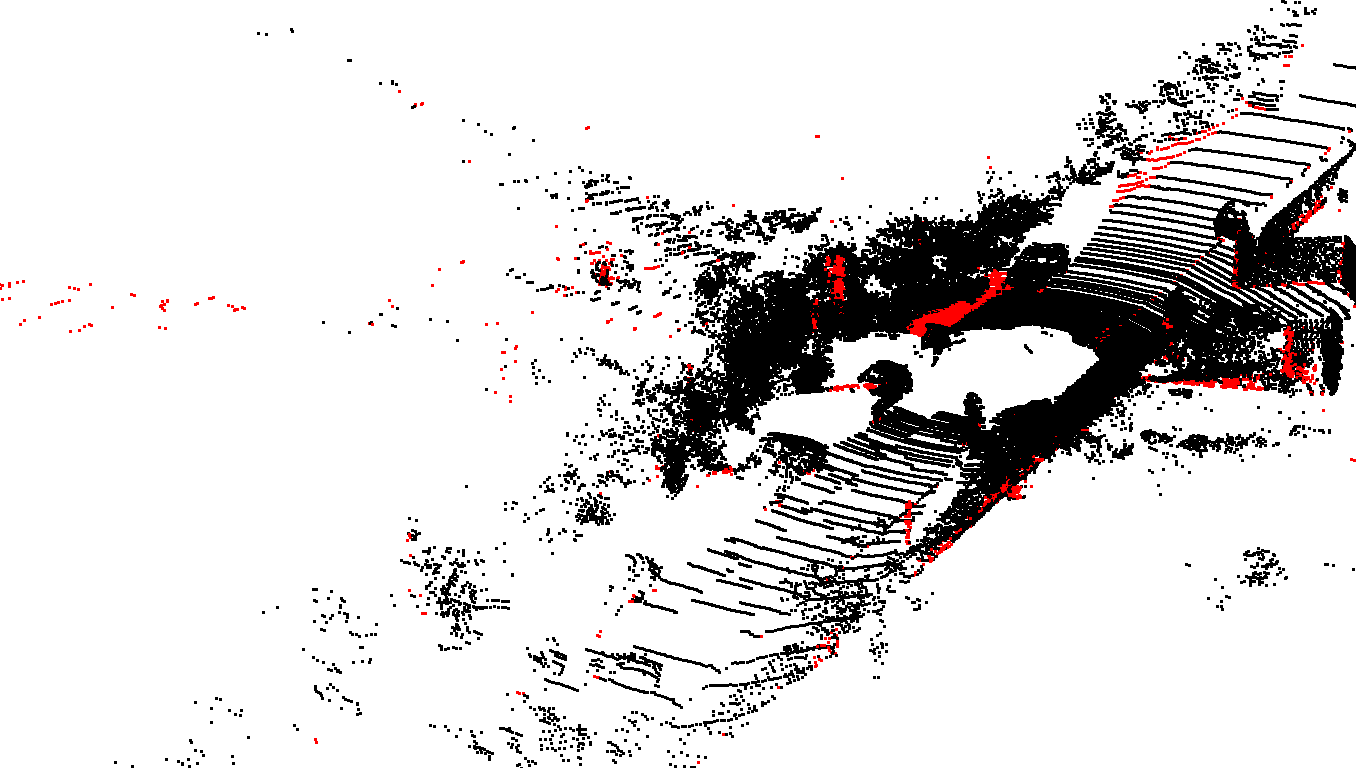}
\\
\caption{Visualization of semantic segmentation results on the validation set of SemanticKITTI obtained with \ours.} 
\label{fig:kitti}
\end{center}
\end{figure*}

\begin{figure*}[h]
\centering
\begin{minipage}{0.75\linewidth}
\centering Color code used for nuScenes data
\end{minipage}
\\
\fbox{\includegraphics[width=.75\linewidth,trim={0cm 0cm 0cm 3cm},clip]{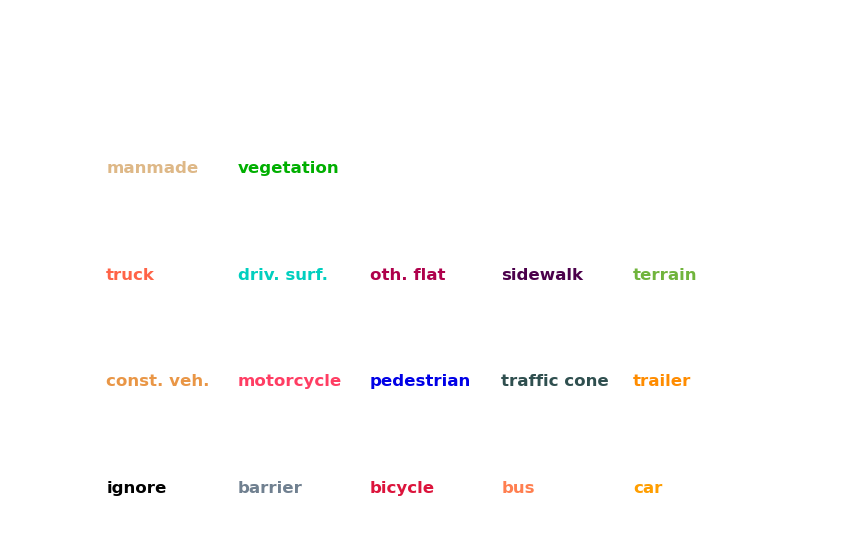}}
\vspace{5mm}
\\
\begin{minipage}{0.75\linewidth}
\centering Color code used for SemanticKITTI data
\end{minipage}
\\
\fbox{\includegraphics[width=.75\linewidth,trim={0cm 0cm 0cm 3cm},clip]{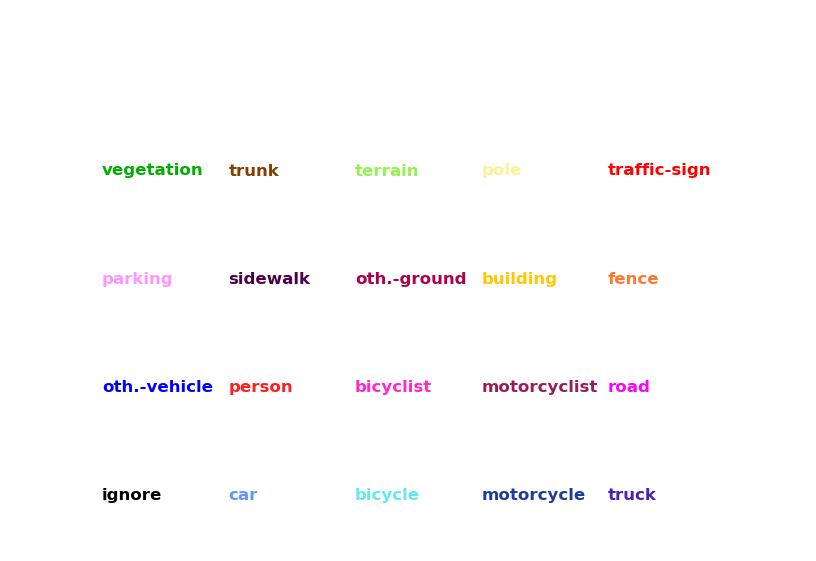}}
\vspace{5mm}
\caption{Color code used to represent each class on nuScenes (top) and SemanticKIITI (bottom).}
\label{fig:color_code}
\end{figure*}

\smallparagraph{Segmentation results.} We present in \cref{fig:nuscenes} and \cref{fig:kitti} visualizations of semantic segmentation results obtained with our method on the validation set of nuScenes \cite{nuscenes} and SemanticKITTI \cite{semkitti}, respectively. The official color codes for these visualizations are recalled in \cref{fig:color_code}.  We notice that, overall, the segmentation are of good quality. Nevertheless, we remark sometimes confusion between the sidewalk and the road on nuScenes (row 1 and 3 in \cref{fig:nuscenes}). We notice as well some wrongly classified points when the vegetation overlaps a building in the last row of \cref{fig:nuscenes}. On SemanticKITTI, we notice essentially some confusion between terrain and vegetation, especially in row 1 and 3 of \cref{fig:kitti}.

\begin{figure*}[h]
\begin{center}
\begin{minipage}{0.04\linewidth}
    \rotatebox{90}{Scene 1}
\end{minipage}
\begin{minipage}{0.95\linewidth}
    \begin{minipage}{0.31\linewidth}
    \centering \small ${\rm Flat}(F^{(0)})$ - $(x, y)$-plane
    \end{minipage}
    \begin{minipage}{0.31\linewidth}
    \centering \small ${\rm Flat}(F^{(24)})$ - $(x, y)$-plane
    \end{minipage}
    \begin{minipage}{0.31\linewidth}
    \centering \small ${\rm Flat}(F^{(42)})$ - $(x, y)$-plane
    \end{minipage}
    \vspace{1mm}
    \\
    \includegraphics[width=0.32\linewidth]{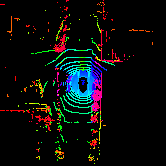}
    \includegraphics[width=0.32\linewidth]{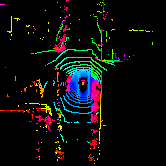}
    \includegraphics[width=0.32\linewidth]{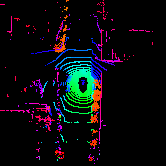}
    \\
    \begin{minipage}{0.31\linewidth}
    \centering \small ${\rm Flat}(F^{(1)})$ - $(x, z)$-plane
    \end{minipage}
    \begin{minipage}{0.31\linewidth}
    \centering \small ${\rm Flat}(F^{(25)})$ - $(x, z)$-plane
    \end{minipage}
    \begin{minipage}{0.31\linewidth}
    \centering \small ${\rm Flat}(F^{(43)})$ - $(x, z)$-plane
    \end{minipage}
    \vspace{1mm}
    \\
    \includegraphics[width=0.32\linewidth,height=1cm]{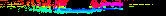}
    \includegraphics[width=0.32\linewidth,height=1cm]{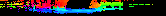}
    \includegraphics[width=0.32\linewidth,height=1cm]{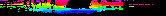}
\end{minipage}
\vspace{1mm}
\\
\rule{\linewidth}{.2mm}
\\
\vspace{1mm}
\begin{minipage}{0.04\linewidth}
    \rotatebox{90}{Scene 2}
\end{minipage}
\begin{minipage}{0.95\linewidth}
    \begin{minipage}{0.31\linewidth}
    \centering \small ${\rm Flat}(F^{(0)})$ - $(x, y)$-plane
    \end{minipage}
    \begin{minipage}{0.31\linewidth}
    \centering \small ${\rm Flat}(F^{(24)})$ - $(x, y)$-plane
    \end{minipage}
    \begin{minipage}{0.31\linewidth}
    \centering \small ${\rm Flat}(F^{(42)})$ - $(x, y)$-plane
    \end{minipage}
    \vspace{1mm}
    \\
    \includegraphics[width=0.32\linewidth]{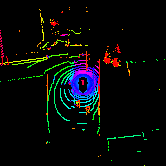}
    \includegraphics[width=0.32\linewidth]{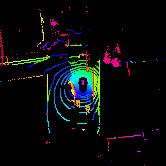}
    \includegraphics[width=0.32\linewidth]{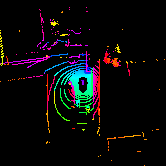}
    \\
    \begin{minipage}{0.31\linewidth}
    \centering \small ${\rm Flat}(F^{(1)})$ - $(x, z)$-plane
    \end{minipage}
    \begin{minipage}{0.31\linewidth}
    \centering \small ${\rm Flat}(F^{(25)})$ - $(x, z)$-plane
    \end{minipage}
    \begin{minipage}{0.31\linewidth}
    \centering \small ${\rm Flat}(F^{(43)})$ - $(x, z)$-plane
    \end{minipage}
    \vspace{1mm}
    \\
    \includegraphics[width=0.32\linewidth,height=1cm]{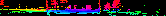}
    \includegraphics[width=0.32\linewidth,height=1cm]{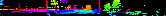}
    \includegraphics[width=0.32\linewidth,height=1cm]{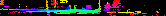}
\end{minipage}
\end{center}
\caption{Visualization of 2D features maps obtained after the Flatten step at different layers  $\ell$ of \ours on two scenes of the validation set of nuScenes. The feature maps are colored by reducing the $F$-dimensional features to a $3$-dimenional space using t-SNE.} 
\label{fig:features_nuscenes}
\end{figure*}

\begin{figure*}[h]
\begin{center}
\begin{minipage}{0.04\linewidth}
    \rotatebox{90}{Scene 1}
\end{minipage}
\begin{minipage}{0.95\linewidth}
    \begin{minipage}{0.31\linewidth}
    \centering \small ${\rm Flat}(F^{(0)})$ - $(x, y)$-plane
    \end{minipage}
    \begin{minipage}{0.31\linewidth}
    \centering \small ${\rm Flat}(F^{(24)})$ - $(x, y)$-plane
    \end{minipage}
    \begin{minipage}{0.31\linewidth}
    \centering \small ${\rm Flat}(F^{(42)})$ - $(x, y)$-plane
    \end{minipage}
    \vspace{1mm}
    \\
    \includegraphics[width=0.32\linewidth]{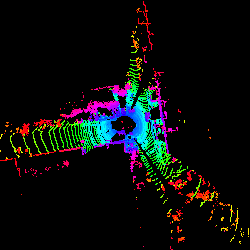}
    \includegraphics[width=0.32\linewidth]{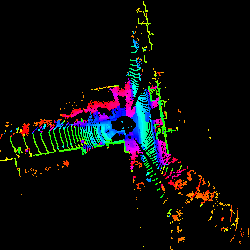}
    \includegraphics[width=0.32\linewidth]{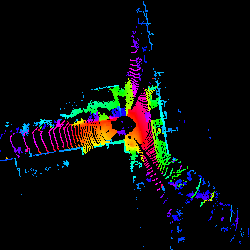}
    \\
    \begin{minipage}{0.31\linewidth}
    \centering \small ${\rm Flat}(F^{(1)})$ - $(x, z)$-plane
    \end{minipage}
    \begin{minipage}{0.31\linewidth}
    \centering \small ${\rm Flat}(F^{(25)})$ - $(x, z)$-plane
    \end{minipage}
    \begin{minipage}{0.31\linewidth}
    \centering \small ${\rm Flat}(F^{(43)})$ - $(x, z)$-plane
    \end{minipage}
    \vspace{1mm}
    \\
    \includegraphics[width=0.32\linewidth,height=5mm]{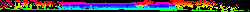}
    \includegraphics[width=0.32\linewidth,height=5mm]{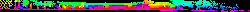}
    \includegraphics[width=0.32\linewidth,height=5mm]{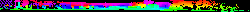}
\end{minipage}
\vspace{1mm}
\\
\rule{\linewidth}{.2mm}
\\
\vspace{1mm}
\begin{minipage}{0.04\linewidth}
    \rotatebox{90}{Scene 2}
\end{minipage}
\begin{minipage}{0.95\linewidth}
    \begin{minipage}{0.31\linewidth}
    \centering \small ${\rm Flat}(F^{(0)})$ - $(x, y)$-plane
    \end{minipage}
    \begin{minipage}{0.31\linewidth}
    \centering \small ${\rm Flat}(F^{(24)})$ - $(x, y)$-plane
    \end{minipage}
    \begin{minipage}{0.31\linewidth}
    \centering \small ${\rm Flat}(F^{(42)})$ - $(x, y)$-plane
    \end{minipage}
    \vspace{1mm}
    \\
    \includegraphics[width=0.32\linewidth]{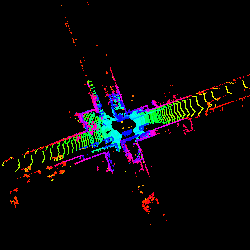}
    \includegraphics[width=0.32\linewidth]{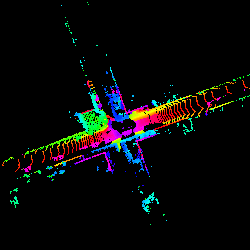}
    \includegraphics[width=0.32\linewidth]{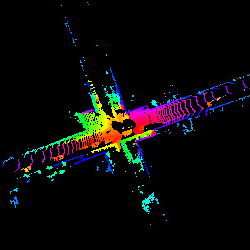}
    \\
    \begin{minipage}{0.31\linewidth}
    \centering \small ${\rm Flat}(F^{(1)})$ - $(x, z)$-plane
    \end{minipage}
    \begin{minipage}{0.31\linewidth}
    \centering \small ${\rm Flat}(F^{(25)})$ - $(x, z)$-plane
    \end{minipage}
    \begin{minipage}{0.31\linewidth}
    \centering \small ${\rm Flat}(F^{(43)})$ - $(x, z)$-plane
    \end{minipage}
    \vspace{1mm}
    \\
    \includegraphics[width=0.32\linewidth,height=5mm]{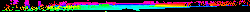}
    \includegraphics[width=0.32\linewidth,height=5mm]{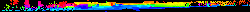}
    \includegraphics[width=0.32\linewidth,height=5mm]{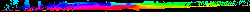}
\end{minipage}
\end{center}
\caption{Visualization of 2D features maps obtained after projection at different layers  $\ell$ of \ours on two scenes of the validation set of SemanticKITTI. The feature maps are colored by reducing the $F$-dimensional features to a $3$-dimensional space using t-SNE.}
\label{fig:features_kitti}
\end{figure*}

\smallparagraph{2D features maps.} For illustration purposes, we provide in \cref{fig:features_nuscenes} and \cref{fig:features_kitti} visualizations of 2D feature maps obtained after projection at different layers $\ell$ of \ours, for nuScenes and SemanticKITTI, respectively.

%
\section{Number of Parameters and Inference time}

The largest WaffleIron models that we trained in this work, \ours-48-256 and \ours-48-384, contain only 6.8 M and 15.1 M trainable parameters, respectively. This stays smaller than, e.g., Cylinder3D \cite{Cylinder3d}, which has more than 50~M parameters.

The inference time for \ours-48-256 is provided in the core of the paper. This inference time does not include data pre-processing. Yet, the only noticeable extra step required in our method for data pre-processing, compared to networks using sparse convolutions, is the nearest neighbor search required to compute the point tokens in the embedding layer. Note that this embedding layer is not tied to our proposed backbone \ours; one could design other embedding layers not requiring this nearest neighbor search.

\medskip

In order to further accelerate inference while keeping the simplicity of implementation of \ours, we can think of the following possibilities which we leave for future work.
\begin{itemize}
    \item Reduce the number of point tokens by increasing the voxel size used for voxel-downsampling during pre-processing. We used square voxels of size $10$ cm, while the 2D grids in the $\ourlayer$ blocks have a resolution of $60$\,cm on nuScenes, and $40$\,cm on SemanticKITTI. We can probably downsample the point clouds further during pre-processing with limited impact on the performance.
    \item Construct a new embedding layer that outputs a reduced number of point tokens, especially in regions that are highly sampled by the lidar and that contain redundant information.
\end{itemize}

\begin{table*}
    
\begin{lstlisting}[language=Python, caption={\textbf{Pytorch implementation of \ours}. This backbone takes as input point tokens provided by an embedding layer, and outputs updated point tokens used in a linear classification layer for semantic segmentation. The implementation of the embedding layer and the classification layer are not presented here. The code to construct the sparse projection matrices on line 57 is presented in \cref{code:spmat}.}, label=code:waffleiron]
import torch
import numpy as np
import torch.nn as nn


class ChannelMix(nn.Module):
    def __init__(self, channels):
        super().__init__()
        # Number of channels denoted by F in the paper
        F = channels 
        # Layers in channel mixing step
        self.norm = nn.BatchNorm1d(F)
        self.mlp = nn.Sequential(nn.Conv1d(F, F, 1), nn.ReLU(), nn.Conv1d(F, F, 1))
        self.layerscale = nn.Conv1d(F, F, 1, bias=False, groups=F) 

    def forward(self, tokens):
        return tokens + self.layerscale(self.mlp(self.norm(tokens)))


class TokenMix(nn.Module):
    def __init__(self, channels, grid_shape):
        super().__init__()
        # Shape of 2D grid on projection plane
        self.H, self.W = grid_shape 
        # Number of channels denoted by F in the paper
        F = channels
        # Layers in token mixing step
        self.norm = nn.BatchNorm1d(F)
        self.ffn = nn.Sequential(
            nn.Conv2d(F, F, 3, padding=1, groups=F), nn.ReLU(), nn.Conv2d(F, F, 3, padding=1, groups=F)
        )
        self.layerscale = nn.Conv1d(F, F, 1, bias=False, groups=F)

    def forward(self, tokens, sp_mat):
        B, C, N = tokens.shape
        # Flatten
        residual = torch.bmm(sp_mat["flatten"], self.norm(tokens).transpose(1, 2)).transpose(1, 2)
        # FFN with channel-wise 2D convolutions with kernels of size 3 x 3
        residual = self.ffn(residual.reshape(B, C, self.H, self.W)).reshape(B, C, self.H * self.W)
        # Inflate
        residual = torch.bmm(sp_mat["inflate"], residual.transpose(1, 2)).transpose(1, 2)
        return tokens + self.layerscale(residual.reshape(B, C, N))


class WaffleIron(nn.Module):
    def __init__(self, channels, depth, grids_shape):
        super().__init__()
        self.grids_shape = grids_shape
        self.channel_mix = nn.ModuleList([ChannelMix(channels) for _ in range(depth)])
        self.token_mix = nn.ModuleList(
            [TokenMix(channels, grids_shape[l % len(grids_shape)]) for l in range(depth)]
        )

    def forward(self, tokens, non_zeros_ind):
        # Build projection matrices
        sp_mat  = [build_proj_matrix(ind, tokens.shape[0], np.prod(sh)) 
                   for ind, sh in zip(non_zeros_ind, self.grids_shape)]
        # Forward pass in backbone
        for l, (smix, cmix) in enumerate(zip(self.token_mix, self.channel_mix)):
            tokens = smix(tokens, sp_mat[l % len(sp_mat)])
            tokens = cmix(tokens)
        return tokens
\end{lstlisting}
\end{table*}
\begin{table*}[!h]
    
\begin{lstlisting}[language=Python, caption={Code to construct the sparse projection matrices used in \ours. Note that we build two matrices for efficiency: one for the Flatten step (`flatten') and one for the Inflate step (`inflate'). The matrix `flatten' combines (i) projection to 2D and (ii) averaging in each 2D cell, i.e., implements Eq.~(4) directly. The matrix `inflate' corresponds to $\ma{S}$ in Eq.~(5).}, label=code:spmat]
def get_non_zeros_ind(point_coord, plane_axes, grid_shape, fov_xyz_min, resolution):
    """ 
    Mapping between point indices and 2D cell indices for the projection from 3D to 2D.
    Inputs:
      `point_coord': xyz-coordinates of the points to project (array of size num_points x 3).
      `planes_axes': axis encoding of projection planes, e.g., `planes_axes=(0,1)' for the (x,y)-plane.
      `grid_shape': shape of 2D grid on projection plane, e.g., `grid_shape=(128,128)'.
      `fov_xyz_min': lowest xyz-bounds of the FOV (array of size 1 x 3)
      `resolution': resolution of 2D grid (scalar)
    Output:
      indices of non-zero entries in sparse matrix for the projection from 3D to 2D.
    """
    
    # Quantize point cloud coordinates at desired resolution
    quant = ((point_coord - fov_xyz_min)[:, plane_axes] / resolution).astype('int')
    # Transform quantized coordinates to 2D cell indices
    cell_indices = quant[:, 0] * grid_shape[1] + quant[:, 1]

    # Indices of non-zeros entries in sparse matrix for projection from 3D to 2D.
    num_points = quant.shape[0]
    indices_non_zeros = torch.cat([
        # Batch index (batch size of 1 here)
        torch.zeros(1, num_points).long(),
        # Index of corresponding 2D cell for each point
        torch.from_numpy(cell_indices).long().reshape(1, num_points),
        # Index of each point
        torch.arange(num_points).long().reshape(1, num_points)
    ], axis=0)

    return indices_non_zeros

    
def build_proj_matrix(indices_non_zeros, batch_size, num_2d_cells):
    """
    Construct sparse matrices for the projection from 3D to 2D and vice versa.
    Inputs:
      `indices_non_zeros': indices of non-zero entries in sparse matrix for projecting from 3D to 2D.
      `batch_size':  batch size.
      `num_2d_cells': number of cells in the 2D grid.
    Outputs:
      sparse projection matrices for the Flatten and Inflate steps.
    """
    num_points = indices_non_zeros.shape[1]
    matrix_shape = (batch_size, num_2d_cells, num_points)

    # One non-zero coefficient per point (set to 1) in sparse matrix for inflate step
    ones = torch.ones(batch_size, num_points, 1, device=indices_non_zeros.device)

    # Sparse projection matrix for Inflate step
    inflate = torch.sparse_coo_tensor(indices_non_zeros, ones.reshape(-1), matrix_shape)
    inflate = inflate.transpose(1, 2)

    # Count number of points in each cells (used in Flatten step)
    num_points_per_cells = torch.bmm(inflate, torch.bmm(inflate.transpose(1, 2), ones))

    # Sparse projection matrix for Flatten step (projection & average in each 2d cells)
    weight_per_point = 1. / num_points_per_cells.reshape(-1)
    flatten = torch.sparse_coo_tensor(indices_non_zeros, weight_per_point, matrix_shape)

    return {"flatten": flatten, "inflate": inflate}
\end{lstlisting}
\end{table*}

\clearpage

\clearpage

\clearpage

\end{document}